%% file: main.tex
\documentclass[nohyperref]{article}

\usepackage{microtype}
\usepackage{graphicx}
\usepackage{subfigure}
\usepackage{booktabs} % for professional tables

\usepackage{hyperref}

\usepackage[accepted]{icml2022}

\usepackage{amsmath}
\usepackage{amssymb}
\usepackage{mathtools}
\usepackage{amsthm}

\usepackage[capitalize,noabbrev]{cleveref}

\input{extra_packages}

\theoremstyle{plain}

\theoremstyle{definition}

\theoremstyle{remark}


\usepackage[textsize=tiny]{todonotes}

\icmltitlerunning{Blocks Assemble! Learning to Assemble with Large-Scale Structured Reinforcement Learning}

\begin{document}

\twocolumn[
% \icmltitle{Learning to Assemble with Large-Scale Reinforcement Learning}
% \icmltitle{Task Assembly Through Large-Scale RL and Structured Agents}
% \icmltitle{Blocks Assemble! \scalerel*{\includegraphics{figures/avengers.png}}{\big(} Learning to Assemble\\with Large-Scale
% % Object-Oriented
% Reinforcement Learning}
% \icmltitle{Blocks Assemble! Learning to Assemble\\with Large-Scale Reinforcement Learning}
% \icmltitle{Learning to Assemble with Large-Scale Structured Reinforcement Learning}
\icmltitle{Blocks Assemble! Learning to Assemble with \\Large-Scale Structured Reinforcement Learning}
% \icmltitle{Open-Ended Assembly through Continual Play}

% It is OKAY to include author information, even for blind
% submissions: the style file will automatically remove it for you
% unless you've provided the [accepted] option to the icml2022
% package.

% List of affiliations: The first argument should be a (short)
% identifier you will use later to specify author affiliations
% Academic affiliations should list Department, University, City, Region, Country
% Industry affiliations should list Company, City, Region, Country

% You can specify symbols, otherwise they are numbered in order.
% Ideally, you should not use this facility. Affiliations will be numbered
% in order of appearance and this is the preferred way.
\icmlsetsymbol{equal}{*}

% \begin{icmlauthorlist}
% % \icmlauthor{Firstname1 Lastname1}{equal,yyy}
% % \icmlauthor{Firstname2 Lastname2}{equal,yyy,comp}
% % \icmlauthor{Firstname3 Lastname3}{comp}
% % \icmlauthor{Firstname4 Lastname4}{sch}
% % \icmlauthor{Firstname5 Lastname5}{yyy}
% % \icmlauthor{Firstname6 Lastname6}{sch,yyy,comp}
% % \icmlauthor{Firstname7 Lastname7}{comp}
% % %\icmlauthor{}{sch}
% % \icmlauthor{Firstname8 Lastname8}{sch}
% % \icmlauthor{Firstname8 Lastname8}{yyy,comp}
% % %\icmlauthor{}{sch}
% % %\icmlauthor{}{sch}

% \icmlauthor{Seyed Kamyar Seyed Ghasemipour}{google}
% \icmlauthor{Daniel Freeman}{google}
% \icmlauthor{Byron David}{google}
% \icmlauthor{Shixiang Shane Gu}{google}
% \icmlauthor{Satoshi Kataoka}{eqsup, google}
% \icmlauthor{Igor Mordatch}{eqsup, google}
% \end{icmlauthorlist}

% \icmlaffiliation{yyy}{Department of XXX, University of YYY, Location, Country}
% \icmlaffiliation{comp}{Company Name, Location, Country}
% \icmlaffiliation{sch}{School of ZZZ, Institute of WWW, Location, Country}

% \icmlaffiliation{google}{Google Research}
% \icmlaffiliation{eqsup}{Co-Supervising Author}

% \icmlcorrespondingauthor{Firstname1 Lastname1}{first1.last1@xxx.edu}
% \icmlcorrespondingauthor{Firstname2 Lastname2}{first2.last2@www.uk}

\begin{center}
   \textbf{
   {\small \color{gray} Seyed} Kamyar {\small \color{gray} Seyed} Ghasemipour
   }$^{1}$, \textbf{Daniel Freeman}$^{1}$, \textbf{Byron David}$^{1}$, \textbf{Shixiang Shane Gu}$^{1}$,\\
   \textbf{Satoshi Kataoka}$^{1}$, \textbf{Igor Mordatch}$^{1}$\\
    $^1$Google Research\\
    \texttt{\{kamyar, cdfreeman, byrondavid, shanegu, satok, imordatch\}@google.com}
\end{center}

% You may provide any keywords that you
% find helpful for describing your paper; these are used to populate
% the "keywords" metadata in the PDF but will not be shown in the document
\icmlkeywords{Machine Learning, ICML}

\vskip 0.3in

% this must go after the closing bracket ] following \twocolumn[ ...

% This command actually creates the footnote in the first column
% listing the affiliations and the copyright notice.
% The command takes one argument, which is text to display at the start of the footnote.
% The \icmlEqualContribution command is standard text for equal contribution.
% Remove it (just {}) if you do not need this facility.

%\printAffiliationsAndNotice{}  % leave blank if no need to mention equal contribution
\printAffiliationsAndNotice{\icmlEqualContribution} % otherwise use the standard text.

\begin{abstract}
Assembly of multi-part physical structures is both a valuable end product for autonomous robotics, as well as a valuable diagnostic task for open-ended training of embodied intelligent agents. We introduce a naturalistic physics-based environment with a set of connectable magnet blocks inspired by children’s toy kits. The objective is to assemble blocks into a succession of target blueprints. Despite the simplicity of this objective, the compositional nature of building diverse blueprints from a set of blocks leads to an explosion of complexity in structures that agents encounter. Furthermore, assembly stresses agents' multi-step planning, physical reasoning, and bimanual coordination. We find that the combination of large-scale reinforcement learning and graph-based policies
-- surprisingly without any additional complexity --
is an effective recipe for training agents that not only generalize to complex unseen blueprints in a zero-shot manner, but even operate in a reset-free setting without being trained to do so. Through extensive experiments, we highlight the importance of large-scale training, structured representations, contributions of multi-task vs. single-task learning, as well as the effects of curriculums, and discuss qualitative behaviors of trained agents.
% \footnote{
% Videos of our results can be viewed at project webpage
Our accompanying project webpage can be found at:
\href{https://sites.google.com/view/learning-direct-assembly/home}{sites.google.com/view/learning-direct-assembly}
% }

% Robotic assembly of objects from a given set of parts is an intrinsically valuable task that touches upon many fruitful avenues of research in Machine Learning. In this work we develop a simulated rendition of this problem in which, given a set of blocks and a blueprint for a desired structure, agents must learn to magnetically assemble the blocks to create the desired form. This simulated benchmark enables us to study many challenges that make assembly a hard problem. Building desired blueprints stresses agents' abilities in physical reasoning, bimanual coordination, and planning the order of blocks to connect. Furthermore, the compositional nature of building diverse blueprints from a set of parts leads to a combinatorial explosion of complexity in structures that agents encounter. Our results demonstrate that the combination of large-scale reinforcement learning (RL) and structured neural network policies (graph neural networks) is an effective recipe which leads to agents that can generalize to complex unseen blueprints in a zero-shot manner, and even operate in a reset-free setting without being trained to do so. Through extensive experiments, we highlight the importance of structured representations, the contributions of multi-task vs. single-task learning, and the effect of using curriculums, and we discuss emergent behaviors of trained agents. We hope that our challenging benchmark can serve as a valuable precursor task for studying the challenges of robotic assembly.

{\centering
\includegraphics[width=\linewidth]{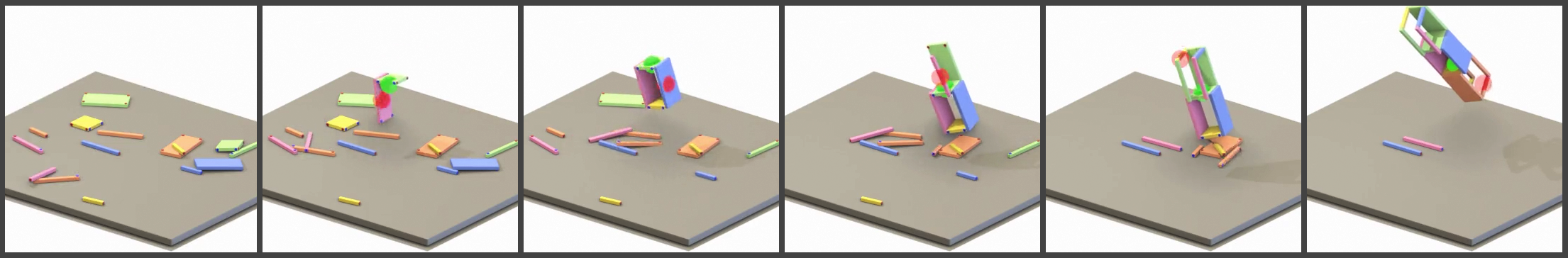}
\par
\label{fig:reset_free}
}

% \begin{center}\par\includegraphics[width=\linewidth]{figures/progress_short.png}\end{center}

\end{abstract}

]

% % Acknowledgements should only appear in the accepted version.
% \section*{Acknowledgements}

% \textbf{Do not} include acknowledgements in the initial version of
% the paper submitted for blind review.

% If a paper is accepted, the final camera-ready version can (and
% probably should) include acknowledgements. In this case, please
% place such acknowledgements in an unnumbered section at the
% end of the paper. Typically, this will include thanks to reviewers
% who gave useful comments, to colleagues who contributed to the ideas,
% and to funding agencies and corporate sponsors that provided financial
% support.

\input{introduction}
\input{assembly_task}
\input{method}
\input{experiments}
\input{related_work}
\input{discussion_and_future}
% \input{ethics}

% \newpage
\input{contributions}
\bibliography{references}
\bibliographystyle{icml2022}

%%%%%%%%%%%%%%%%%%%%%%%%%%%%%%%%%%%%%%%%%%%%%%%%%%%%%%%%%%%%%%%%%%%%%%%%%%%%%%%
%%%%%%%%%%%%%%%%%%%%%%%%%%%%%%%%%%%%%%%%%%%%%%%%%%%%%%%%%%%%%%%%%%%%%%%%%%%%%%%
% APPENDIX
%%%%%%%%%%%%%%%%%%%%%%%%%%%%%%%%%%%%%%%%%%%%%%%%%%%%%%%%%%%%%%%%%%%%%%%%%%%%%%%
%%%%%%%%%%%%%%%%%%%%%%%%%%%%%%%%%%%%%%%%%%%%%%%%%%%%%%%%%%%%%%%%%%%%%%%%%%%%%%%
\newpage
\appendix
\onecolumn
\input{appendix}

% You can have as much text here as you want. The main body must be at most $8$ pages long.
% For the final version, one more page can be added.
% If you want, you can use an appendix like this one, even using the one-column format.
%%%%%%%%%%%%%%%%%%%%%%%%%%%%%%%%%%%%%%%%%%%%%%%%%%%%%%%%%%%%%%%%%%%%%%%%%%%%%%%
%%%%%%%%%%%%%%%%%%%%%%%%%%%%%%%%%%%%%%%%%%%%%%%%%%%%%%%%%%%%%%%%%%%%%%%%%%%%%%%

\end{document}

%% file: extra_packages.tex
\usepackage{scalerel}
\usepackage{listings}
\usepackage{multirow}

\usepackage{xcolor}

\definecolor{codegreen}{rgb}{0,0.6,0}
\definecolor{codegray}{rgb}{0.5,0.5,0.5}
\definecolor{codepurple}{rgb}{0.58,0,0.82}
\definecolor{backcolour}{rgb}{0.95,0.95,0.92}

\lstdefinestyle{mystyle}{
    backgroundcolor=\color{backcolour},   
    commentstyle=\color{codegreen},
    keywordstyle=\color{magenta},
    numberstyle=\tiny\color{codegray},
    stringstyle=\color{codepurple},
    basicstyle=\ttfamily\footnotesize,
    breakatwhitespace=false,         
    breaklines=true,                 
    captionpos=b,                    
    keepspaces=true,                 
    numbers=left,                    
    numbersep=5pt,                  
    showspaces=false,                
    showstringspaces=false,
    showtabs=false,                  
    tabsize=2
}

%% file: introduction.tex
\section{Introduction}
    % \begin{figure}[t]
    %     \centering
    %     \includegraphics[width=0.8\linewidth]{figures/env.pdf}
    %     \caption{An example of the environments we consider.}
    %     \label{fig:env}
    % \end{figure}
    \begin{figure*}[ht]
        \centering
        \includegraphics[width=1.0\linewidth]{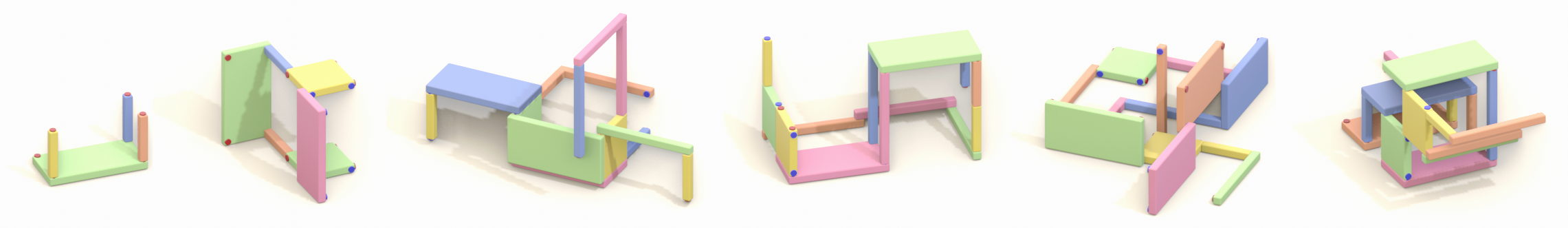}
        \caption{Examples of target blueprints we consider. We train on variety of target structures, ranging from structures of 2 to 16 blocks.}
        \label{fig:blueprints}
    \end{figure*}
    Robotic assembly of objects from a given set of parts is an incredibly intriguing avenue for research in artificial intelligence (AI). Not only is it a valuable capability we would like autonomous robots to possess, but it is a challenging problem statement with open-ended complexity, touching upon many fruitful avenues of AI research. Agents that can assemble structures can reshape their surroundings, which creates dynamic environments with more possibilities for open-ended learning ~\citep{baker2019emergent,wang2019paired,co2020ecological}.
    % an intrinsically valuable end-goal with significant opportunity for positive societal and economic impact, but it is a challenging problem statement leading to many fruitful avenues of Machine Learning research.
    A key feature of assembly is that due to its compositional and modular nature, given a set of parts, one can create objects on a broad spectrum of complexity. Furthermore, in order to solve the assembly problem, agents must acquire a diverse set of skills and capabilities. They must learn to grasp and attach components in an order that is amenable to successful completion. They must develop physical reasoning capabilities to avoid collision, and they need to learn bi-manual coordination. In addition, once agents have acquired such skills, it is expected that they can rapidly learn to create new desired objects.
    
    % The first key contribution of our work is to 
    To study assembly, we
    create a simulated but naturalistic environment which consists of blocks of varying shapes that can be magnetically attached to one-another. Agents are then tasked with constructing a desired structure from one of almost 200 pre-designed blueprints (although generative models can be used to automate this process ~\citep{thompson2020building}). In this environment, in lieu of grasping robotic arms, we enable agents to directly move desired blocks. Such direct manipulation and the use of magnetic connections (instead of more complex joining mechanisms) abstracts away some details of the assembly problem while retaining many of the challenges that make the problem hard and interesting. Compared robot manipulation tasks which emphasize rearrangement and stacking ~\citep{li2020towards,openai2021asymmetric}, the compositional nature of assembly from a set of blocks leads to an agent continually encountering new structures of varying complexities, which necessitates developing a richer representation of what constitutes an ``object" ~\citep{spelke1990principles}. In addition, our magnetic assembly benchmark stresses multi-step planning, physical reasoning, and bimanual coordination.
    
    % \kamyar{Despite difficulty, can find a single policy that solves everything, hammer home the key contributions here. }
    Despite the complexity of this problem, we find that it is possible to train a single agent that can simultaneously assemble all given blueprint tasks, generalize to complex unseen blueprints in a zero-shot manner, and even operate in a reset-free manner despite being trained in an episodic fashion. Our solution relies on a combination of large-scale reinforcement learning, structured (graph-based) agent representations, and simultaneous multi-blueprint training. Simultaneously training on diverse blueprints of varying complexity scaffolds the agent's learning by enabling it to first make progress of simpler tasks (such as simply joining two blocks), while structured policy representations enable the agent to generalize and transfer its solutions towards solving more complex -- and even unseen -- blueprints.
    % the agent to solve simple tasks first (such as simply joining two blocks). Structured policy representations allow solutions to these simple tasks to be transferable and scaffold learning of subsequently more complex blueprints.
    We empirically observe a progression of learning increasingly large blueprints - many of which were not solvable with single-blueprint training.
    Surprisingly, we find other components such as planning or hierarchical approaches to be unnecessary for this task.
    Through experiments, we highlight the contributions of various components such as structured policies, episodic initial state distribution, curriculum that emphasize training on harder blueprints, and discuss qualitative behaviors and maneuvers discovered by the trained agents.
    
    % Towards presenting a solution our magnetic assembly benchmark, we study how traditional RL methods fare on this task, and discover that large-scale on-policy RL with graph structured policies (graph neural networks\kamyar{CITATIONS?}) leads to very effective policies that can  This is a particularly surprisingly result given that intuitively we expected hierarchical reinforcement learning or planning 
    % \kamyar{CITATIONS} to be a key ingredient in the solution to the assembly tasks. Our second surprising finding is that multi-task training is an essential ingredient in enabling agents to assemble complex structures. 
    
    Our contributions are as follows:
    \begin{itemize}
    \item We introduce an assembly domain that allows for a controlled study of generalization in reinforcement learning (RL).
    \item We demonstrate a single agent that can simultaneously solve all seen assembly tasks and generalize to unseen tasks.
    \item We demonstrate the importance of combining large-scale RL, structured policies, and multi-task training as a route to arrive at generally capable agents.
    \end{itemize}
    % We will open-source the environment and training code and policy checkpoints and
    We hope this work further encourages study of assembly as an open-ended means to develop and evaluate embodied agent learning.

%% file: assembly_task.tex
\section{Magnetic Block Assembly Environment}
    \begin{figure}
        \centering
        \includegraphics[width=0.9\linewidth]{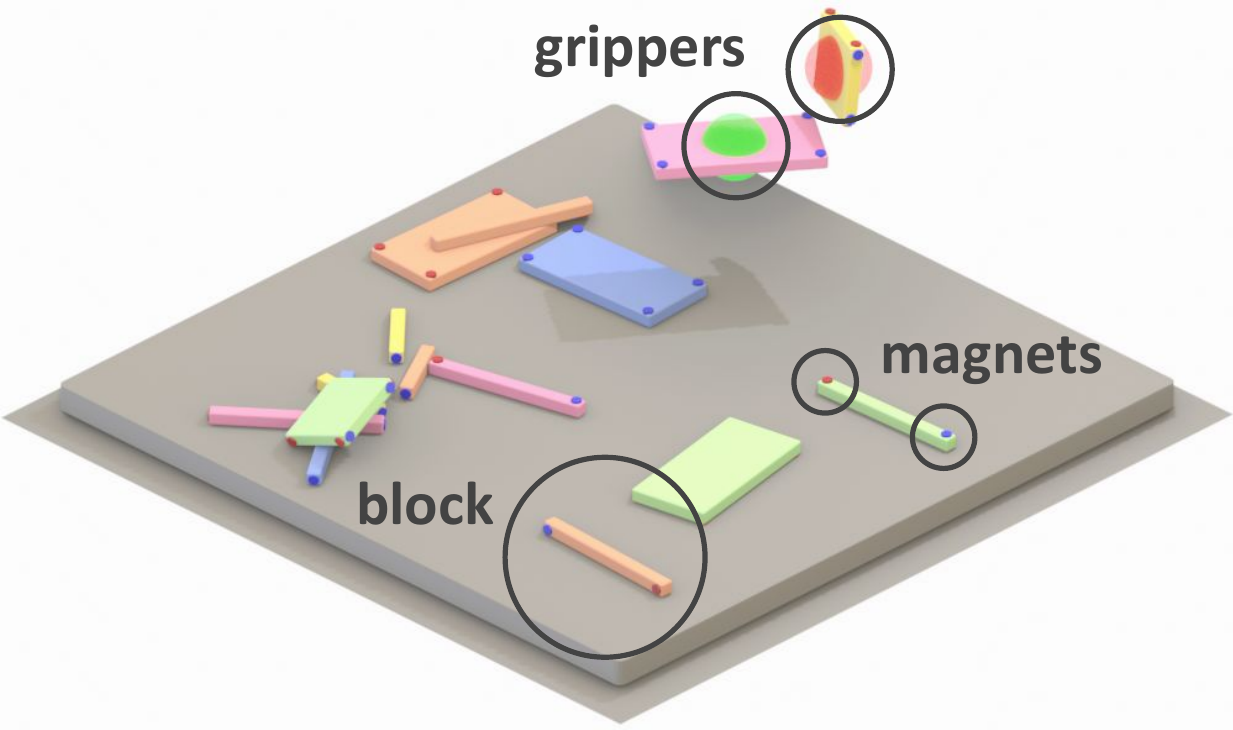}
        \caption{Our magnetic assembly domain. Two virtual grippers directly manipulate the available blocks to magnetically assemble a desired blueprint structure.}
        \label{fig:env}
    \end{figure}
    % \kamyar{Read Shane's comment.}
    Our goal is to design a minimal tractable assembly environment to study generalization in a naturalistic, multi-step, combinatorial, dynamic problem requiring bi-hand coordination.    
    We construct a three-dimensional environment containing a fixed set of 16 cuboid blocks of 6 different types. Blocks contain positive and negative magnet points, rendered as red and blue respectively, positioned on the block surface. Positive and negative magnets ``snap" together when sufficiently close, and disconnect when adequate pulling force is applied. Magnets enable creation of arbitrarily complex composed structures from the given building blocks.
    % \kamyar{I don't understand this next sentence, why can't other connection types be implemented in the real world? Also sentence needs a bit of rewriting.} \im{tried to clarify - better? I was thinking about instant weld constraints like deepmind construction paper or hide and seek tractor beams} 
    Additionally, magnets can be implemented in the real-world, unlike more abstract locking constraints (e.g. instantaneous weld constraints), yet are tolerant enough to join objects without tackling the problem of high-precision insertion that would be required for other connection mechanisms such as pegs or screws. To simplify the problem, in lieu of robotic arms we opt for the use of virtual grippers which can directly manipulate desired blocks. More specifically, each gripper can decide which block to move, and set its positional and rotational velocities. The use of direct manipulation abstracts away the challenges of grasping and manipulation with a robotic arm, and enables us to focus on research questions concerning higher-level assembly behaviors such as planning and generalization to unseen structures.
    % \shane{(optional) you can cite some prior work making similar assumption in control research, e.g. team2021open abstracts object pick up with direct control as well i believe }.
    While the number of grippers is parameterizable, unless otherwise specified, throughout this work we will use 2 virtual grippers.
    % \im{Say a bit about how direct grippers abstract away the hard picking and manipulation problem and let us focus on higher-level assembly behavior.}
    
    To specify the assembly task, we designed 165 blueprints (split into 141 train, 24 test) describing interesting structures to be built, although the blueprints can potentially be procedurally generated~\citep{thompson2020building}. The complexity of the created blueprints range from requiring only a single magnetic connection, up to challenging structures that make use of all 16 available blocks. The problem statement in our magnetic assembly environment is simple to describe: In each episode, the agent must assemble the blocks to create the desired blueprint. Each episode begins with either all blocks randomly scattered around the environment, or from a randomly sampled pre-constructed blueprint -- with unused blocks dispersed on the ground. Episodes are 100 environment steps long, translating to a length of 10 seconds in the real world.
    %with each environment step advancing the physics simulation by 10 steps.
    In each step agents receive rewards based on how close blocks are to their intended configurations, as well as correct and incorrect magnetic connections. Episodes terminate when exactly correct magnetic connections are made and blueprint blocks are in the correct relative position and orientation, or when 100 steps has passed.
    Our simulated assembly task is implemented in the open-source Mujoco \citep{todorov2012mujoco} physics engine. A detailed description of observation space, action space, rewards, and success criterion used can be found in Appendix \ref{app:env_description}.

% \begin{figure*}[t]
%     \centering
%     \includegraphics[width=0.9\linewidth]{figures/blueprints-wide.png}
%     \caption{Examples of target blueprints we consider. We train on variety of target structures, ranging from structures of 2 to 16 blocks.}
%     \label{fig:blueprints}
% \end{figure*}

%% file: method.tex
\section{Methodology}
    While solving the magnetic assembly task can be approached through a variety of solutions such as hierarchical reinforcement learning and geometric planning algorithms (e.g. RRT \citep{lavalle2001rapidly}), in past work it has been demonstrated that many tasks which intuitively require complex planning strategies can be solved through large-scale application of reinforcement learning algorithms using the right training setups, and appropriate neural network architectures and inductive biases ~\citep{silver2017mastering,berner2019dota,vinyals2019grandmaster,baker2019emergent}. Motivated by such results, in this work we explore the ingredients necessary to train effective agents for magnetic assembly through RL. In the subsequent sections we describe the main ingredients of training successful agents and study the contribution of each component.
    
\section{Agents}
    \label{sec:agents}
    In this section we describe our structured agents, including observations and action spaces, and graph-based network architecture. The Python code describing our agent architecture can be found in Appendix \ref{app:architecture}.
    % IGOR: moved to training section
    %s which are implemented using Jax~\citep{jax2018github}, Jraph~\citep{jraph2020github}, Haiku~\citep{haiku2020github}, and Acme~\citep{hoffman2020acme} libraries.
    
    % \kamyar{We have tried different architectures that were more expensive to run, but went with this one because it was much faster, and there wasn't much gain it seemed (maybe we didn't train long enough?).}\df{VERY briefly mention what didn't work, or possibly kick to an appendix, but do mention here.}

    % \begin{figure}
    %     \centering
    %     \includegraphics[width=\linewidth]{figures/architecture.pdf}
    %     \caption{\im{Give details about this figure.}}
    %     \label{fig:architecture}
    % \end{figure}
    \begin{figure}[t]
        \centering
        \includegraphics[width=\linewidth]{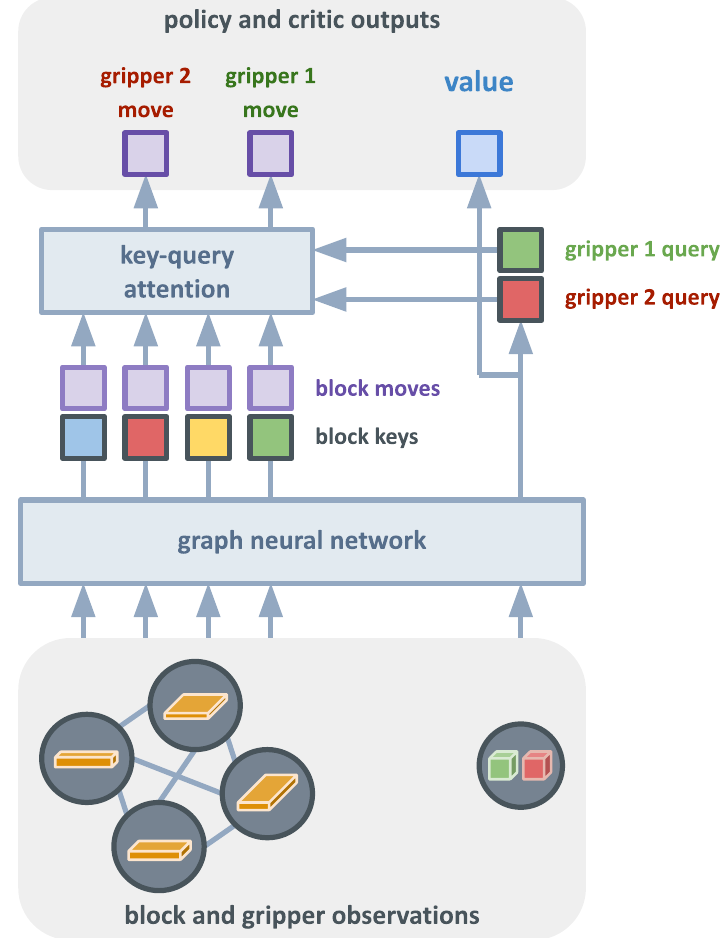}
        \caption{
            \small{
                Diagram depicting our structured agent. Inputs to the agent are graph-structured block observations as well as gripper observations. A graph neural network processes the observations and produces: (1) per block moves and per block attention keys, (2) per gripper attention queries, (3) a global latent representation. Using dot-product attention between the keys and queries, the grippers decide which block to hold, and output its proposed move. The global latent representation is used to predict a baseline value for the PPO~\citep{schulman2017proximal} algorithm.
            }
        }
        \label{fig:architecture}
    \end{figure}
    
    \subsection{Structured Observations}
        The observations provided to the agent can be divided into two broad categories, those concerning the blocks, and those concerning the grippers. When designing observations to provide to agents, we have taken the effort to ensure observations are invariant to the global position and orientation. This is a valuable inductive bias that provides agents with the flexibility build desired blueprint anywhere and in any rotation.
        
        \paragraph{Block Observations}
        In the assembly task, observations pertaining to the blocks can be naturally organized into a directed graph, with each node containing information about a particular block, and each directed edge representing relative information about the two blocks. The information contained in each node is very minimal: the \texttt{z} height of the block from the ground, and whether it was being held by each gripper in the previous timestep. The majority of observations are placed on the directed edges. An edge connecting two blocks contains the information regarding: relative position and orientation of their magnets that need to be connected, change in relative position and orientation of the blocks needed to match the blueprint, relative position of center of mass of the blocks, whether the blocks are magnetically attached, and whether the blocks should be magnetically attached according to the blueprint. All these observations can be automatically extracted from the simulator state and the target blueprint configuration, and can realistically be computed in a real-world setting as well by simply obtaining each blocks position and orientation.
        Detailed information regarding exact observations can be found in Appendix \ref{app:env_description}.
        
        \paragraph{Gripper Observations}
        For each gripper we include its orientation, positional and rotational velocities, and which block the gripper was holding in the previous timestep.
    
    \subsection{Graph Neural Network Encoder}
        Given that our magnetic assembly task can be naturally set up using graph-based observations, prior to extracting actions and critic values, we first encode inputs using a graph neural network architecture~\citep{battaglia2018relational}, specifically graph attention networks \citep{velivckovic2017graph}.
        The two inputs to our encoder are (1) a directed graph containing all block observations (2) a ``global node" containing gripper observations. After linearly embedding all input features, they are passed through $N=3$ graph attention layers whose design is inspired by Transformers~\citep{vaswani2017attention} and Graph Attention Networks~\citep{velivckovic2017graph}. Concisely, in each layer, each node aggregates information by attending to incoming edge and node features, and subsequently the global node features are updated by aggregating information from the graph nodes.
        An intuitive diagram describing the architecture used can bee seen in Figure \ref{fig:architecture}.
        % A detailed description of the neural network architecture used can also be found in the source code included in Appendix \ref{app:architecture}.
    
    \subsection{Policy}
        Through experimentation, we have discovered that in addition to a graph neural network encoder, a key design choice is how to extract policy actions from the encoded inputs.
        The outputs of the graph neural network encoder are hidden features per node corresponding to the blocks, and hidden features corresponding to the global node. Using linear layers, from each block node we obtain 2 vectors: (1) a vector representing how the block would like to be moved if a gripper chooses to move it, and (2) a vector representing a key vector for the block.
        % , which will be used to decide which gripper moves which block.
        From the global hidden features, using linear layers we obtain one query vector per gripper. To obtain logits representing which gripper decides to move which block, we use dot-product attention between the block keys and the gripper query vectors, akin to the popular attention mechanism used in Transformers \citep{vaswani2017attention}. In addition to the use of graph neural networks, using this form of decoding from the graph encoder has been a key enabler in training effective policies.
    
    \subsection{Critic}
        To train our RL agents we also require critic value estimates, which we obtain by passing global features obtained from the graph encoder to a 3 layer MLP, with 512 dimensional hidden layers, and relu activation function.

\section{Training and Evaluation}
    \label{sec:training}
    \paragraph{Large-Scale PPO}
    We train our agents using Proximal Policy Optimization (PPO) ~\citep{schulman2017proximal} and Generalized Advantage Estimation (GAE) ~\citep{schulman2015high}, and follow the practical PPO training advice of ~\citep{andrychowicz2020matters}. As will be shown below, one of the most key ingredient in enabling the training of our magnetic assembly agents is the scale of training. Unless otherwise specified, our agents are trained for 1 Billion environment timesteps, using 1 Nvidia V100 GPU for training, and 3000 preemptible CPUs for generating rollouts in the environment. 1 Billion steps in our setup amounts to about 48 hours of training. The key libraries used for training are Jax~\citep{jax2018github}, Jraph~\citep{jraph2020github}, Haiku~\citep{haiku2020github}, and Acme~\citep{hoffman2020acme}.
    % IGOR: skip this due to time?
    % \kamyar{Can we generate a dollar estimate based on GCP prices for how expensive it would be to train one of these agents?}
    
    \paragraph{Multi-Task Training}
    The blueprints that we have designed range from very simple 2 block structures, up to complex blueprints containing all blocks. To train assembly agents, we have split blueprints into training and testing structures, and unless otherwise specified, agents are trained on the full training set of blueprints; in each episode, we sample a training blueprint and task the agent with creating that structure.

    \paragraph{Initial State}
    Episodes start from either (1) all the blocks randomly dispersed on the ground, or (2) a randomly chosen preconstructed blueprint structure with unused blocks randomly arranged on the ground. Resetting from blueprints increases the diversity of initial states, forces the agent to learn how to disassemble structures, and as we found enables a reset-free mode of operation where the agent can continually construct, deconstruct, and reconstruct different blueprints. Unless otherwise specified, we reset from training blueprints with probability 0.2.
    
    \paragraph{Curriculum}
    We have observed that throughout training, some blueprints can be quickly learned while others can be much more challenging. We believe an interesting feature of the assembly problem is that even if a blueprint is currently unsolvable, due to the modular aspect of building complex structures, agents can learn more effectively if we emphasize focus on more challenging blueprints rather than allocating resources to rolling out policies on blueprints that they are already capable of solving. For this reason, in each episode we sample goal blueprints based on a curriculum whose detailed description can be found in Appendix \ref{app:curriculum}.

    \paragraph{Performance Evaluation}
    During training, we evaluate trained policies continuously approximately every 10 minutes by freezing the policy and computing 
    average success rate over
    % average episode return, average success rate, and average steps to succeed of 
    40 episodes.  This continuous evaluation is executed on both training and test environments.
    % in parallel on dedicated CPUs.
    Also, in each evaluation cycle, we generate a video to visualize the agents' behavior.
    Such visualizations have been a valuable asset in iterating over the design of our agents, observations, reward functions, and training setups.
    % This generates a better understanding of training performance for different types of features (e.g. block sizes, magnet connections, maximum height).

%% file: experiments.tex
\section{Experiments}
    \subsection{Importance of Large-Scale Training}
        \begin{figure*}[t]
            \centering
            \includegraphics[width=\linewidth]{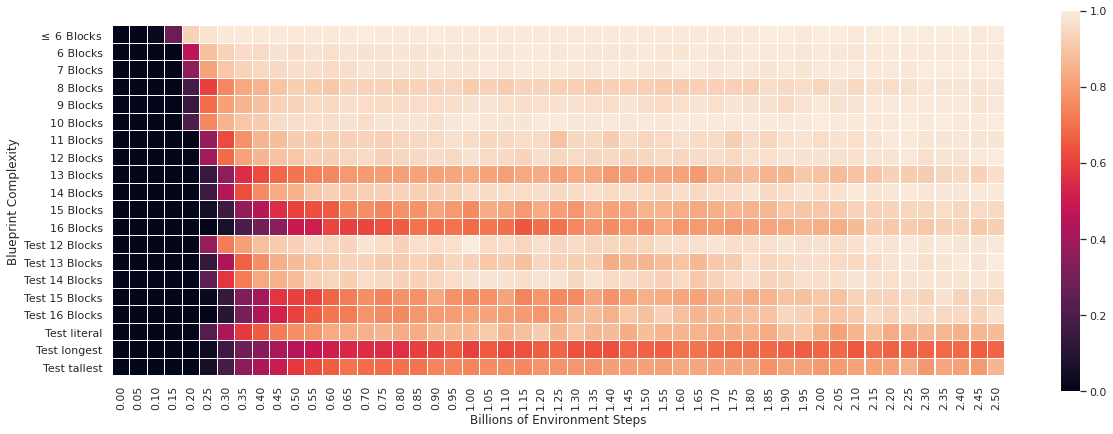}
            \caption{
            Plot presenting success rates for different groups of blueprints throughout training. Each square represents the success rate on 40 episodes, evaluated at that point in training, averaged across two training runs. Our results demonstrate (1) the key role of scale in training successful agents, and that (2) after a long period of training, agents generalize well to complex held-out blueprints.
            Per blueprint results presented in Figure \ref{fig:super_policy_full}.
            }
            \label{fig:super_policy}
        \end{figure*}
        
        We begin by verifying that our training procedure leads to capable assembly agents. To this end, we train our structured agents (Section \ref{sec:agents}), using the training procedure described in Section \ref{sec:training}, for 2.5 billion environment steps to observe training patterns that may arise over a long period of training. Figure \ref{fig:super_policy} presents the success rates of our agent (averaged across two runs) throughout training, on blueprints the agent was trained on as well as held-out structures (per blueprint success rates presented in Figure \ref{fig:super_policy_full} in the appendix).
        
        The first key observation is the compute scale necessary for effectively training our structured agents using PPO \citep{schulman2017proximal}. The simplest 2 block structures can take up to 100 million steps to be reliably solved, while it can take up to 500 million environment steps until the first time some of the most complex blueprints are solved. The second observation is that after a long period of training, not only can agents reliably solve all training blueprints, but they can also generalize well to complex held-out blueprints.

    \subsection{Multi-Task vs. Single-Task}
        \begin{table*}[h]
        \centering
        \begin{tabular}{|l||cc|cc|}
        \hline
        \multirow{2}{*}{\textbf{Blueprint Size}} & \multicolumn{2}{c|}{\textbf{Single-Blueprint Training}}               & \multicolumn{2}{c|}{\textbf{Multi-Blueprint Training (ours)}}                \\ \cline{2-5} 
                                   & \multicolumn{1}{c|}{\textbf{Success Rate}} & \textbf{Steps Until Success} & \multicolumn{1}{c|}{\textbf{Success Rate}} & \textbf{Steps Until Success} \\ \hline
        6 Blocks          & \multicolumn{1}{c|}{99.6\%}       & 100M                & \multicolumn{1}{c|}{100\%}        & 180M                \\ \hline
        12 Blocks         & \multicolumn{1}{c|}{99.9\%}       & 480M                & \multicolumn{1}{c|}{98.8\%}       & 220M                \\ \hline
        16 Blocks         & \multicolumn{1}{c|}{0\%}          & --                  & \multicolumn{1}{c|}{90.9\%}       & 240M                \\ \hline
        \end{tabular}
        \caption{Comparison of Single-Blueprint vs. Multi-Blueprint training for blueprints of various complexities. Success rates are calculated after 1 billion steps of training for the respective agents, and ``Steps Until Success" denotes approximately the first timestep at which the respective agent was able to successfully create the blueprint. While for very small structures single-blueprint training can be effective, for more complex blueprints, single-blueprint agents take significantly longer or are entirely unable to learn the task within 1 billion timesteps.}
        \label{tab:single_vs_multi}
        \end{table*}
        As noted in Section \ref{sec:training}, agents are simultaneously trained to construct all blueprints in the training split. To understand the contribution of this ``Multi-Task" training, we train three agents in a ``Single-Task" setting: one for learning to construct a particular 6 block blueprint, one for constructing a particular 12 block blueprint, and one for constructing a particular 16 block blueprint. The success rates for these three agents can be found in Figures \ref{fig:just_6_005}, \ref{fig:just_12_000}, and \ref{fig:just_16_001} respectively. Our key observations are the following: (1) While the 6 and 12 block blueprints are eventually learned, the 16 block blueprint is not learned, (2) In the single-task setting, the 12 block blueprint requires approximately 500 million environment steps to be learned, while in the multi-task setting (Figure \ref{fig:super_policy_full}) it is learned within 300 million steps, (3) the single-task agents can transfer to some blueprints of equal or lower complexity than they were trained on, but mostly fail to transfer to any blueprints they were not trained to solve. This is in sharp contrast to the multi-task agents which can even transfer to complex held-out blueprints. These results highlight the necessity of multi-task training, not only for generalization to unseen blueprints, but for quickly and reliably solving complex tasks, despite the fact that agent architectures are well-matched to the problem domain.

    \subsection{Structured Agent Architecture}
        \label{sec:structured_agents_exp}
        \begin{table*}[t]
        \centering
        \small{
        \begin{tabular}{|l||ccccc|}
        \hline
        \multirow{2}{*}{\textbf{Agent Type}} & \multicolumn{5}{c|}{\textbf{Success Rate}}                                                                                                                                                        \\ \cline{2-6} 
                                    & \multicolumn{1}{c|}{\textbf{Train 2 - 6 Blocks}} & \multicolumn{1}{c|}{\textbf{Train 7 - 11 Blocks}} & \multicolumn{1}{c|}{\textbf{Train 12 - 16 Blocks}} & \multicolumn{1}{c|}{\textbf{Test 12 - 16 Blocks}} & \textbf{Test Special} \\ \hline
        \textbf{Default}               & \multicolumn{1}{c|}{\textbf{99.9\%}}             & \multicolumn{1}{c|}{\textbf{97.9\%}}              & \multicolumn{1}{c|}{\textbf{87.3\%}}               & \multicolumn{1}{c|}{\textbf{89.6\%}}              & \textbf{77.5\%}       \\ \hline
        \textbf{No Graph Attention}     & \multicolumn{1}{c|}{95.3\%}             & \multicolumn{1}{c|}{43.1\%}              & \multicolumn{1}{c|}{4.1\%}                & \multicolumn{1}{c|}{3.6\%}               & 4.8\%        \\ \hline
        \textbf{Non-Graph Network}      & \multicolumn{1}{c|}{0\%}                & \multicolumn{1}{c|}{0\%}                 & \multicolumn{1}{c|}{0\%}                  & \multicolumn{1}{c|}{0\%}                 & 0\%          \\ \hline
        \textbf{Single-Gripper}        & \multicolumn{1}{c|}{94.6\%}             & \multicolumn{1}{c|}{77.9\%}              & \multicolumn{1}{c|}{51.7\%}               & \multicolumn{1}{c|}{55.4\%}              & 42.8\%       \\ \hline
        \end{tabular}
        }
        \caption{Comparing the success rates of agent ablations, discussed in Sections \ref{sec:structured_agents_exp} and \ref{sec:bimanual}, on the various subsets of blueprints. Results are reported after 1 billion steps of training the respective agents.}
        \label{tab:agent_comparisons}
        \end{table*}
        As discussed in section \ref{sec:agents}, given that state information for the assembly task can be naturally organized into a graph representation, the use of graph neural networks imbues agents with an inductive bias that is well-matched to the domain. Indeed, prior work \citep{bapst2019structured, li2020towards} have observed that the use of agents with relational structures is a key ingredient in solving object-oriented tasks. In this section we aim to understand the contribution of various components of our agents' architecture.
        
        \paragraph{Removing Attention in Graph Layers}
        Figure \ref{fig:no_attention_summary} demonstrates the effect of removing the attention mechanism in the graph neural network layers, meaning that while our agents continue have a graph inductive bias, the hidden representations for each block are updated by treating other blocks equally, rather than deciding which blocks to attend to. The results in Table \ref{tab:agent_comparisons} and Figure \ref{fig:no_attention_summary} clearly demonstrate the necessity of the attention mechanism.
        
        \paragraph{Removing Relational Inductive Biases}
        We also attempt to train agents without relational inductive biases. Instead, we flatten the environment observations and use a residual network \citep{he2016deep} encoder, with a similar action and value decoder as described in Section \ref{sec:agents}. Details of the residual network architecture are described in Appendix \ref{app:resnet}. We train three variants of the residual network agents: (1) trained on the full training set of blueprints, (2) trained on a subset of the training blueprints requiring $\leq 6$ blocks, and (3) trained on a single training blueprint requiring $6$ blocks. Figures \ref{fig:resnet_all}, \ref{fig:resnet_full}, and \ref{fig:resnet_6_005} demonstrate that in all three scenarios, removing the relational inductive bias of graph neural networks is catastrophic. After 1 billion steps, all agents have a $0\%$ success rate on all blueprints, and never accomplish higher than $2.5\%$ success rate on any train or held-out blueprint.
    
    \subsection{Bimanual Manipulation}
        \label{sec:bimanual}
        Compared to most robotics tasks and benchmarks~\citep{levine2018learning,kalashnikov2018qt,andrychowicz2020learning,chen2021system,huang2021generalization,yu2020meta,li2020towards,batra2020rearrangement,openai2021asymmetric}, part-based assembly stresses the bimanual coordination of agents. To verify that our magnetic assembly domain stresses this skill, we compare success rates between bimanual and single-gripper agents in Table \ref{tab:agent_comparisons} and Figure \ref{fig:bimanual_summary}. We find that while our single-gripper agents finds unique strategies to complete some of the structures, its overall success rate is lower than that of a dual-gripper agent, particularly on the more complex blueprints. This indicates the necessity of using two grippers in our domain.
        % \df{no mention of 3+ grippers?  Even just a reference with maybe one figure in the appendix would be a neat teaser, here}
    
    \subsection{Reset-Free Evaluation}
        % \begin{figure*}[t]
        %     \centering
        %     \includegraphics[width=\linewidth]{figures/progress_short.png}
        %     \caption{Still frames showing progression of assembly. \im{best not to duplicate this figure. Let's keep it in the teaser but remove it here?} }
        %     \label{fig:reset_free}
        % \end{figure*}
        As described in Section \ref{sec:training}, with probability $0.2$, the initial state of an episode is set to be a randomly selected pre-constructed blueprint, with remaining blocks dispersed on the ground. This choice has two advantages: (1) it provides an opportunity for agents to learn how to disassemble incorrect constructions, and (2) it enables the evaluation of our agents in a reset-free manner, where we continually task agents with constructing new blueprints without resetting the environment to an initial state.
        
        To evaluate the contribution of this choice, we compare the success rate of two agents, one with and one without blueprint resets, in a reset-free setting. Specifically, within one reset-free episode we ask agents to build 10 consecutive blueprints without resetting to an initial state: once an agent successfully constructs a blueprint, or the maximum of 100 steps has elapsed, we change the target blueprint. As an additional challenge, we sample blueprints from the training set structures requiring a minimum of 12 blocks. We report the success rate aggregated across 50 reset-free episodes (i.e. 50 $\times$ 10 total episodes).
        % and show progression of one such re-assembly in Figure \ref{fig:reset_free}.
        
        When resetting from blueprints is disabled, our agent achieves a sucess rate of $69.4\% \pm 17.0\%$. In constrast, with blueprint resets, the success rate increases to $93.1\% \pm 7.5\%$. This is an exciting finding as it demonstrates a scenario where episodic training enables agents to be deployed in the practically-relevant reset-free scenario.

    \subsection{Curriculum}
        As described in Section \ref{sec:training}, throughout training we make use of a curriculum that increases the likelihood of sampling more challenging blueprints. To analyze its contribution, we compare two runs of agents with curriculum, to an agent trained without curriculum. Our results in Figures \ref{fig:curr_0_minus_no_curr} and \ref{fig:curr_1_minus_no_curr} indicate that our curriculums do not have a clear-cut benefit, but may be leading to improvements in generalization to blueprints in the held-out test set.
        % \df{there's a cite here I think---I believe openai found aggressive randomizing to be about as good as curricula? maybe in the rubix cube work?}.

    \subsection{Curbing Unrealistic Behaviors}
        % \kamyar{Give a language example of an unrealistic behavior. Calculate the rate of block switching before and after fine-tuning. Mention that when using this delay mechanism failures on the test set for tallest and longest increase because they have to keep changing blocks and they run out of time. Discuss rates of block switching.}
        Due to the use of direct manipulation, agents can rapidly switch which block they are holding, which can result in sometimes unrealistic maneuvers not achievable by physical robot grippers. Thus, in the event we wanted to transfer the success from our direct manipulation environment to more realistic settings using robotic arms, it is important to understand how one can mitigate unrealistic behaviors. To this end, after training an agent using the default training procedure described in Section \ref{sec:training}, we modify the environment as follows: Whenever a gripper chooses to change the object it is holding, we disable that gripper for 2 steps. After this change, we continue to train our agent.
        
        Figure \ref{fig:super_minus_grip_trans_delay} demonstrates that while initially the agent's success rate drops very significantly, within less than 100 million environment steps the agent recovers its strong performance. This is a small amount of steps compared to the 2.5+ billion environment steps to train the agent.
        
        Given this result, one might ask whether we could have started training our agents using this environment modification from the beginning. Figure \ref{fig:super_minus_grip_trans_delay_from_scratch} compares this approach to our default training setup. As can be seen, training agents from scratch using gripper transition delays is a significantly more challenging problem, and even after 1.5 billion environment steps, the agent is still unable to make significant progress on many of the blueprints in the training set. These results demonstrate that an efficient approach towards training practical agents is to first train agents in the simplest settings, and continue to finetune those agents in more realistic scenarios. Videos demonstrating behaviors of agents discussed in this section can be found in the accompanying project webpage.
        % \kamyar{Videos demonstrating the behaviors for agents discussed in this section can be found at this link.}\shane{this reminds me of Go-Explore-ish motivation: solve in deterministic setting then robustify. optional to mention it}
    
    \subsection{Analyzing Learned Solutions}
        In this section our goal is to obtain a qualitative understanding of the strategies learned by our trained agents.
        
        \paragraph{Learned Attention Patterns}
        % \kamyar{Maybe aside from videos, there would be a way to show this with pictures in the appendix too?}
        As shown in Section \ref{sec:structured_agents_exp}, the attention mechanism is a key ingredient in the graph neural network architecture we used. To understand what the attention heads in the different layers have learned to focus on, we visualize agents' attention patterns throughout different episodes. We observe the following interpretable patterns: (1) Some attention heads focus on the blocks that should be connected according to the blueprint, (2) Some keep account of which blocks are currently connected, and (3) Some focus on which blocks are currently being held by the two grippers. Other attention heads, particularly in the later layers, are more challenging to interpret, but appear to contain a combination of the previously described attention motifs. Videos visualizing learned attention patterns can be found in the accompanying website.
        
        \paragraph{Qualitative Behaviours}
        Rolling out trained agents, we observe a number of interesting learned behaviors. Examples of such behaviors include: (1) Despite environment observations not providing fine-grained detail about free-space, agents appear to have learned robust collision avoidance skills. (2) When building complex structures, agents appear to first build separate smaller substructures, and subsequently attach the substructures to construct the full blueprint.
            % \begin{itemize}
            %     \item What happens at different points in training
            %     \item evolution of capabilities and behaviors
            % \end{itemize}
        
        % \paragraph{Failure Cases}
        %     \begin{itemize}
        %         \item Compare checkpoints at 1B and 5B.
        %     \end{itemize}

%% file: related_work.tex
\section{Related Work}

\paragraph{Assembly and Construction}
~\citet{bapst2019structured} previously studied two-dimensional construction environments, training an agent to assemble a structure for an open-ended goal, such as a connecting or a covering structure. Certain details of assembly are abstracted away, as the agent has the ability to directly summon a block of choice anywhere in the scene and weld blocks via an explicit action. By contrast, our environment contains a fixed set of blocks that must be moved -- or reassembled from a previous structure -- in three-dimensional space, and where block connections are made via magnet forces. Such a design makes the environment more easily implementable as a real-world robot setup.
~\citet{lee2019ikea,lee2021adversarial} also introduce a three-dimensional assembly environment for furniture design from a blueprint. By contrast, we use generic blocks, which leads to combinatorial complexity in the space of structures, and enables a more controlled study of generalization.
A key point of differentiation from the above works is that we use assembly as a domain for the study of generalization, and train a single agent to solve all -- seen and unseen -- assembly tasks simultaneously. Additionally, we introduce the use of multiple grippers in assembly which allows us to evaluate the bimanual coordination of trained agents.
~\citet{chung2021brick} present a task where given side view images of a desired structure, Lego blocks must be stacked to create a structure with a similar silhouette. At each step, the state of the Lego structure is encoded using graph neural networks, and through deep RL, a policy learns where the next Lego block must be placed. In contrast, in our work we focus on the dynamic task of assembly, where discrete decisions and continuous control are solved simultaneously using large-scale deep RL.
% Closely related to our work is the work of ~\citet{funk2022learn2assemble}.
~\citet{funk2022learn2assemble} introduce a 3D task where blocks must be re-arranged to create stable structures that occupy randomly sampled target regions. They present a long-horizon manipulation algorithm combining deep reinforcement learning and Monte-Carlo tree search, that at each step decides which block should be moved to which location. Similar to our work, multi-head attention graph neural networks are used, which enable generalization to new settings with larger number of blocks. By comparison, we train a single policy to jointly consider the low-level manipulation of the blocks and block selection.
% In comparison, in our work we do not use low-level primitives. Instead, we focus on the more dynamic task of bimanual parts-based assembly, and study how to simultaneously solve planning as well as low-level control using deep reinforcement learning.
~\citet{suarez2018can} studied assembly of a single chair with real-world bimanual robots using offline planning methods. ~\citet{kim2019shallow,cabi2019scaling} studied real-world insertion problem, which is an operation in the broader assembly process.
~\citet{hartmann2021long} present a planning system to solve long-horizon multi-robot construction problems, consisting of stacking parts to create architectural structures. In contrast, in this work we focused on the assembly problem and were motivated to understand whether deep reinforcement learning can be used so simultaneously solve discrete planning and continuous control.

\paragraph{Generalization in Robotic Manipulation}
Much recent work in robot manipulation focused on the tasks of object grasping ~\citep{levine2018learning,kalashnikov2018qt}, in-hand object manipulation ~\citep{andrychowicz2020learning,chen2021system,huang2021generalization}, or execution of a motor skill ~\citep{yu2020meta}, where variation comes from diversity of object shapes and arrangements involved. By contrast, while assembly uses a fixed set of blocks, the compound structures that must be manipulated have a combinatorial diversity of shape that dynamicaly changes during the episode.
~\citet{li2020towards,batra2020rearrangement,openai2021asymmetric} propose scene re-arrangement as a universal task for embodied AI. While re-arrangement is general, we caution that many instances of rearrangement can be solved as a sequence of largely independent sub-tasks that do not influence each other, and can be performed in any order. By contrast, assembly steps are coupled and must be performed in a specific order that must be discovered by the agent. ~\citet{ahmed2020causalworld, funk2022learn2assemble} also present robotics benchmarks based on block rearrangement, with stronger emphasis on long-horizon planning, learning from structured representations, and generalization to unseen scenarios. ~\citet{gupta2019relay} introduce a kitchen environment with an implicit dependency structure among sub-tasks (i.e. opening a container before putting something in it), but such pre-conditions are harder to scale.

\paragraph{Structured and Object-Centric Policies}
Besides our work, there have been numerous efforts to parameterize policies  through structured models such as graph neural networks or Transformers, leveraging intrinsic objectness and invariances of the physical world~\citep{spelke1990principles}. They have been used for controlling agents of different morphologies for locomotion and manipulation tasks~\citep{wang2018nervenet,sanchez2018graph,chen2018hardware,pathak2019learning,huang2020one,kurin2020my}, as well as enabling agents to learn a compositionally-challenging task like stacking~\citep{li2020towards}. Other relevant works include parameterizing manipulation actions, especially of 2D tasks, as object-based spatial actions, which have enabled breakthroughs in vision-based manipulation~\citep{zeng2020transporter,noguchi2021tool,shridhar2022cliport}. While inspired by similar motivations, our work tackles a uniquely challenging task, 3D bi-manual assembly, compared to single-arm or 2D tasks in prior work~\citep{li2020towards,zeng2020transporter}.

%% file: discussion_and_future.tex
\section{Limitations, Discussion, \& Future Directions}

% \kamyar{Say something about not needing planning, doing bimanual, and reset-free}
% \kamyar{Do we need to mention reset-free here?} \sk{Reset-free is especially useful for robust experiments in real but currently the behavior for reset-free is a bit less realistic (in the most cases, the agent flings the structure on the ground :).  So, I think it might be good to mention that there are some space to improve the behavior as one of our limitations.} \im{I think it's fine to skip it, since we discuss real-world realism considerations later and I think people will assume we mean that there}
We introduced a new blueprint assembly environment for studying bimanual assembly of multi-part physical structures, and demonstrated training of a single agent that can simultaneously solve all seen and unseen assembly tasks via a combination of large-scale RL, structured policies, and multi-task training. While our work showed that a solution to our problem exists, it is by no means efficient - requiring billions of training episodes. It is likely that by incorporating planning or hierarchical methods, the training time can be significantly shortened. Additionally, upon maturity of accelerated simulation engines~\citep{liang2018gpu,brax2021github,makoviychuk2021isaac}, our agents may be trained at a similar compute scale using much more modest hardware infrastructures.
Beyond more efficient training, in this work we chose to abstract away complexities of manipulation and perception. A more detailed treatment of these elements, such as constraints that prevent overly aggressive behaviors, can bring this work closer to robotics applications.
% \im{"including constraints that prevent overly aggressive behavior"?}.
And lastly, while we focused on blueprint assembly in this work, more open-ended assembly goals as well as multi-agent interaction present further research opportunities for developing agents of increasing complexity.

    % \begin{itemize}
    %     \item efficiency, HRL, planning
    %     \item Building for tool use, cite structured agents
    % \end{itemize}

%% file: contributions.tex
\section{Author Contributions}
Kamyar Ghasemipour: First-author, ran all experiments, wrote manuscript draft, developed the agent architectures (e.g. Graph Neural Network architectures, policy action representations, etc.), contributed to the environment creation (improved and added observations, rewards, success conditions, moved observations from node-based towards edge-based structure, bug fixes).

Byron David, Daniel Freeman, Shane Gu: Helped experiment implementations, creating blueprints used for creating the magnetic parts assembly, helped with discussing ideas.

Satoshi Kataoka: Co-supervising author, created majority of robust infrastructure (evaluation infrastructure, and backup mechanisms), contributed to the environment creation (improved robustness, usability and reliability), reviewed all code modifications.

Igor Mordatch: Co-supervising author, initiated the research direction, created the environment as well benchmark task definition, created first proof of concept results motivating more focused investigation, guided project direction, rendered figures and videos.

%% file: appendix.tex
\section{Blueprints}
\input{all_blueprints}

\clearpage
\section{Descriptions of Observation Space, Action Space, Rewards, Success Criterion}
\label{app:env_description}
    \subsection{Observation Space}
        \paragraph{Per block observations on graph nodes}
            \begin{itemize}
                \item $z$ height from ground
                \item binary indicator for whether the block was being held in the previous timestep by some gripper
            \end{itemize}
        
        \paragraph{Observations on directed graph edge from block A to block B}
            \begin{itemize}
                \item change in position for desired magnets to align
                \item change in orientation for desired magnets to align
                \item the difference between current block positional delta, and block positional delta in the blueprint
                \item the difference between current block orientation delta, and block orientation delta in the blueprint
                \item positional delta
                \item 0-1 indicator for whether blocks should be connected according to the blueprint
                \item 0-1 indicator for whether blocks should are currently connected
            \end{itemize}
        
        \paragraph{Per gripper observations}
            \begin{itemize}
                \item orientation
                \item positional velocity
                \item rotational velocity
                \item 16-dimensional 0-1 vector showing which block the gripper was holding in the previous timestep
            \end{itemize}
    
    \subsection{Action Space}
        \begin{itemize}
            \item one 16-dimensional one-hot vector per gripper, representing which block the gripper wants to manipulate
            \item a 16 $\times$ 6 dimensional matrix, representing desired position and rotational velocity per block, were it to be chosen by a gripper to be manipulated
        \end{itemize}
    
    \subsection{Rewards}
        In each step, the following rewards are computed and added together. The cumulative reward is then subtracted from the cumulative reward in the prior timestep, and the difference in cumulative rewards is returned as the environment reward for the current timestep.
        \begin{itemize}
            \item force penalty if two blocks, or a block and the ground, make heavy contact
            \item -1 for each magnetic connection that should not be connected
            \item dense reward based on position and orientation of each two magnets that should be connected
            \item +1 if two magnets that should be connected are connected
            \item dense reward based on position and orientation of each two blocks that should be attached to one-another
        \end{itemize}
    
    \subsection{Success Criterion}
        An episode is deemed successful when in some timestep all of the following success criterion are satisfied.
        \begin{itemize}
            \item magnetic connections required by the blueprint are made
            \item no extra magnets outside the blueprint definition are connected
            \item all blocks in the blueprint structure are in the correct relative position and orientation
        \end{itemize}
    
% \kamyar{A detailed description of observation space, action space, rewards, and success criterion used can be found in Appendix \ref{app:env_description}.}

\section{Graph Neural Network Architecture}
\label{app:architecture}
\input{agent_code}

\section{Curriculum}
\label{app:curriculum}
The following python script represents how our training curriculum is defined.

\begin{lstlisting}[language=Python]
import numpy as np
from scipy import special

class Curriculum():
  def __init__(
      self,
      tau: float = 0.2,
      temp: float = 0.5,
      decay: float = 0.99,
      curriculum_logdir: str = ''):
    self.tau = tau
    self.temp = temp
    self.decay = decay
    self.blueprint_indices = blueprint_indices

    self.success_rates = np.zeros((len(blueprint_indices),), dtype=np.float32)

    self.probs = np.ones((len(blueprint_indices),), dtype=np.float32)
    self.probs *= 1./float(len(blueprint_indices))
    self.probs_history = self.probs[None, :].copy()

    self._steps = 0

  def update(self, batch):
    """Updates the curriculum."""
    successes = batch['success']
    blueprint_indices = batch['blueprint_indices']

    for s, b_i in zip(successes, blueprint_indices):
      self.success_rates[i] = (1. - self.tau) * self.success_rates[i] + self.tau * s

    self.success_rates *= self.decay # ensures there are not things that never get sampled
    self.probs = special.softmax((1. - self.success_rates) / self.temp)
  
  def sample(self):
    return int(np.random.choice(np.arange(len(blueprint_indices)), p=self.probs))
\end{lstlisting}

\section{ResNet Baseline Details}
\label{app:resnet}
When ablating the role of relation inductive biases in the agent architecture, we replace the graph neural network encoder in our default agent (described in Section \ref{sec:agents}) with a residual network encoder defined as follows:

\begin{lstlisting}[language=Python]
import jax
import haiku as hk

NUM_RESNET_BLOCKS = 4
HIDDEN_DIM = 1024

uniform_initializer = hk.initializers.VarianceScaling(
    scale=0.333, mode='fan_out', distribution='uniform')

def residual_block(x):
    h = x
    h = hk.Linear(HIDDEN_DIM, w_init=uniform_initializer)(h)
    h = jax.nn.relu(h)
    h = hk.Linear(HIDDEN_DIM, w_init=uniform_initializer)(h)
    x = x + h
    return hk.LayerNorm(axis=-1, create_scale=True, create_offset=True)(x)

def resnet_encoder(x):
    for _ in range(NUM_RESNET_BLOCKS):
          x = block(x)
\end{lstlisting}

\input{full_size_plots}

%% file: all_blueprints.tex
\begin{figure}[h]
    \centering
    \includegraphics[width=0.80\linewidth]{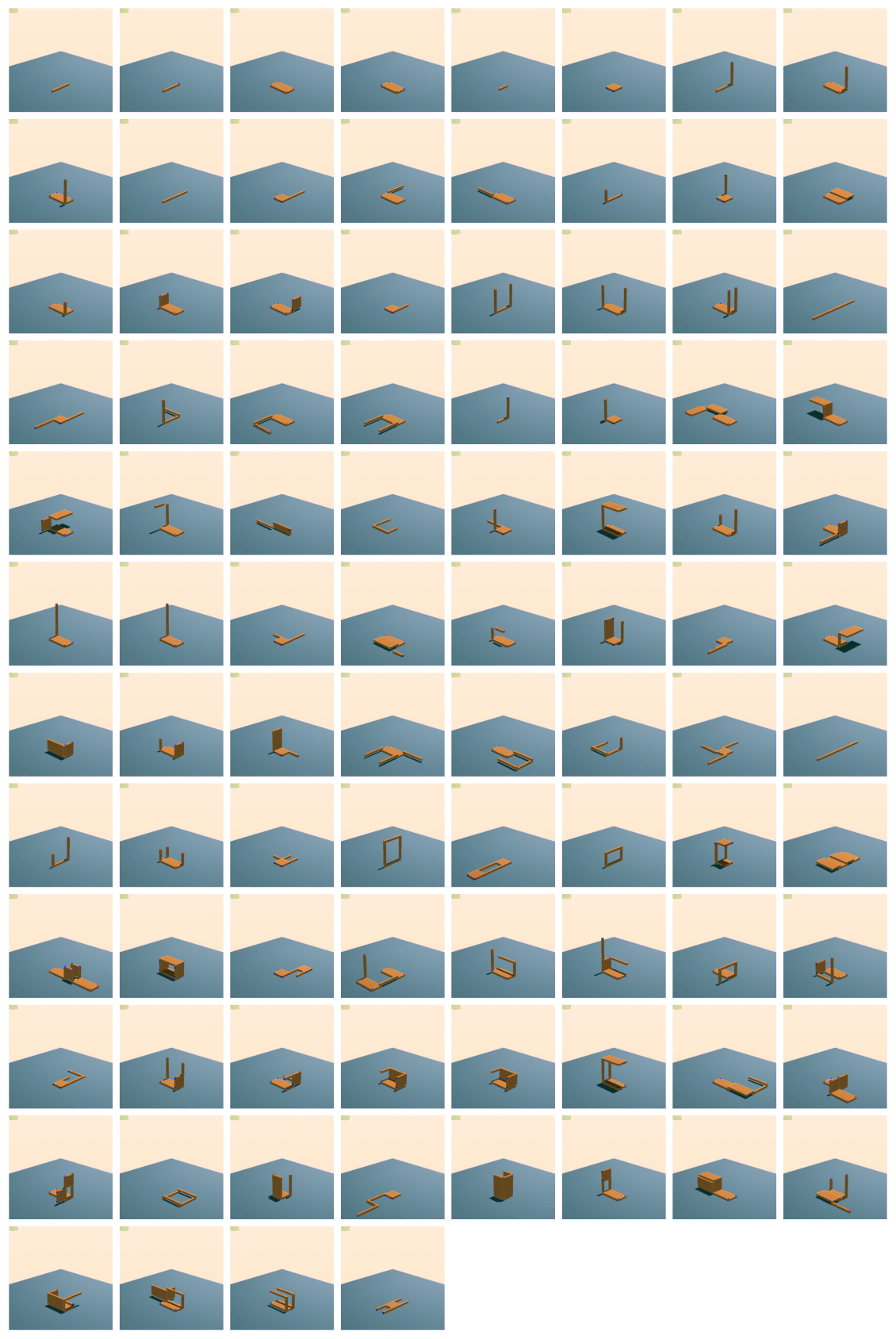}
    \caption{All blueprints used for training (of size 1 - 5).}
    \label{fig:1_5}
\end{figure}
\begin{figure}[h]
    \centering
    \includegraphics[width=0.80\linewidth]{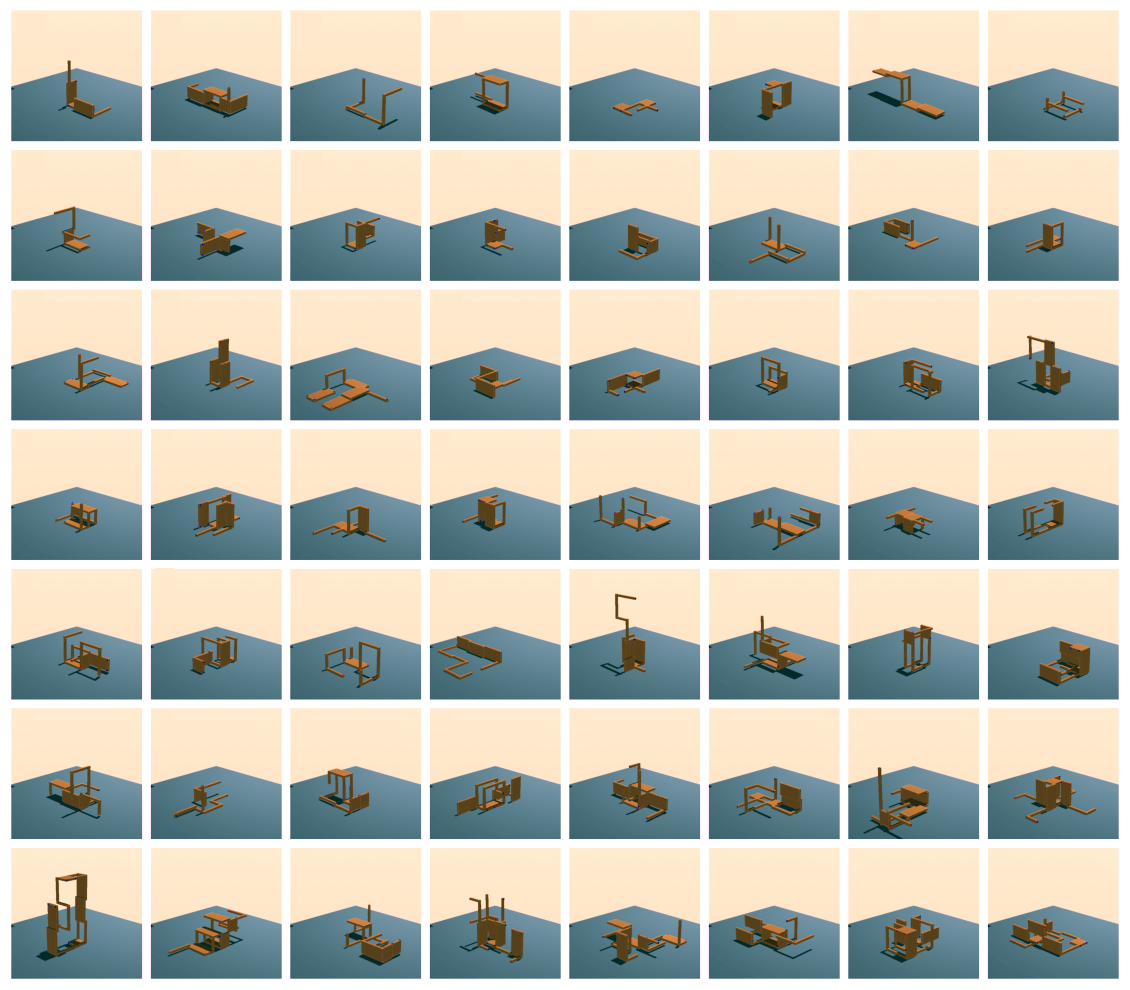}
    \caption{All blueprints used for training (of size 6 - 16).}
    \label{fig:1_6}
\end{figure}
\begin{figure}[h]
    \centering
    \includegraphics[width=0.80\linewidth]{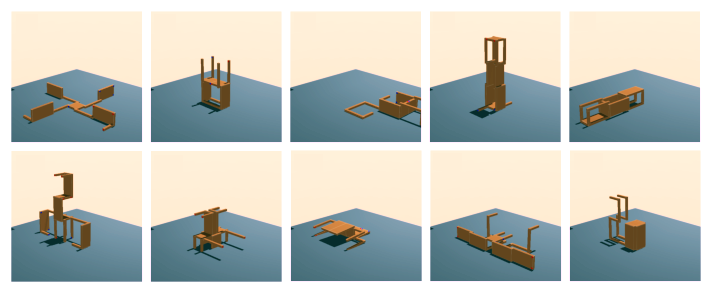}
    \caption{All blueprints used for testing (of size 12 - 16).}
    \label{fig:1_7}
\end{figure}
\begin{figure}[h]
    \centering
    \includegraphics[width=0.80\linewidth]{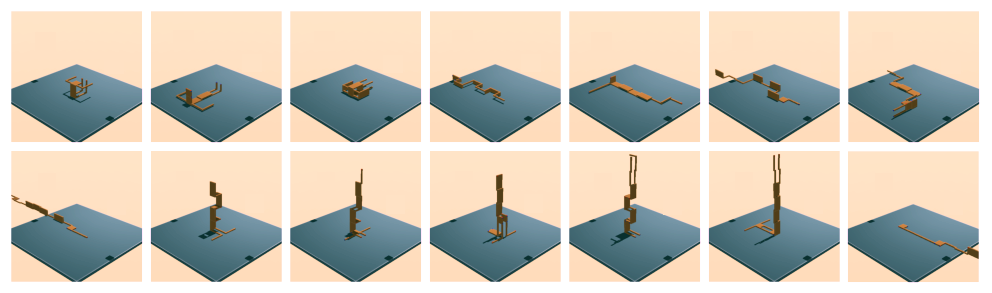}
    \caption{Blueprints used for literal, tallest, longest testing.}
    \label{fig:1_8}
\end{figure}

In this section, we describe details of all blueprints.  Figure \ref{fig:1_5} and \ref{fig:1_6} are used for training while figure \ref{fig:1_7} and \ref{fig:1_8} are used for testing.  Table \ref{tab:2_0} shows distribution for block sizes for different types.


\begin{table}[h]
\centering
\begin{tabular}{|c|c|c|c|c|c|}
\hline
\multicolumn{1}{|c||}{Number of blocks} & \multicolumn{5}{c|}{Types}  \\ \cline{2-6}
\multicolumn{1}{|c||}{} & Training & Test & Test Literal  & Test Longest & Test Tallest \\ \hline
\multicolumn{1}{|c||}{2}  & 14 & & & &  \\
\multicolumn{1}{|c||}{3}  & 31 & & & &  \\
\multicolumn{1}{|c||}{4}  & 36 & & & &  \\
\multicolumn{1}{|c||}{5}  & 5  & & & &  \\
\multicolumn{1}{|c||}{6}  & 9  & & & &  \\
\multicolumn{1}{|c||}{7}  & 6  & & & &  \\
\multicolumn{1}{|c||}{8}  & 6  & & & &  \\
\multicolumn{1}{|c||}{9}  & 6  & & & &  \\
\multicolumn{1}{|c||}{10} & 5  & & 1 & &  \\
\multicolumn{1}{|c||}{11} & 4  & & & &  \\
\multicolumn{1}{|c||}{12} & 4  & 2 & 1 & 1 & 1 \\
\multicolumn{1}{|c||}{13} & 3  & 2 & & 1 & 1 \\
\multicolumn{1}{|c||}{14} & 5  & 2 & 1 & 1 & 1 \\
\multicolumn{1}{|c||}{15} & 4  & 2 & & 1 & 1 \\
\multicolumn{1}{|c||}{16} & 4  & 2 & 1 & 1 & 1 \\ \hline
\end{tabular}
\caption{Number of blueprint block distributions}
\label{tab:2_0}
\end{table}

%% file: agent_code.tex
\begin{lstlisting}[language=Python]
import jax
import jax.numpy as jnp
import haiku as hk
import jraph

def build_GraphNetGAT_ppo_network(environment_spec):
  action_dim = np.prod(environment_spec.actions.shape, dtype=int)

  #### PPO NETWORK PARAMS
  NUM_GN_GAT_LAYERS = 3
  VALUE_MLP_LAYER_SIZES = [512, 512, 512]
  #### GN_GAT LAYER PARAMS
  NUM_HEADS = 4
  NUM_READOUT_HEADS = 4
  # NUM_HEADS = 8
  QK_DIM = 64
  M_DIM = 64
  NODE_EMBED_DIM = 64
  EDGE_EMBED_DIM = 64
  GLOBAL_EMBED_DIM = 64
  NODE_MLP_HID = 256
  NODE_MLP_NUM_LAYERS = 1
  GLOBAL_MLP_HID = 256
  GLOBAL_MLP_NUM_LAYERS = 1
  
  def update_edge_fn(edge_features, sender_features, receiver_features, global_features):
    feats = jnp.concatenate([sender_features, edge_features], axis=-1)
    m = hk.Linear(NUM_HEADS * M_DIM, with_bias=False, name='linear_message')(feats)
    return (jnp.reshape(m, list(m.shape[:-1]) + [NUM_HEADS, M_DIM]), edge_features)

  def update_node_fn(node_features, sender_features, receiver_features, global_features):
    receiver_features = jnp.reshape(receiver_features, list(receiver_features.shape[:-2]) + [-1])
    x = jnp.concatenate([receiver_features], axis=-1)
    residual = hk.Linear(node_features.shape[-1], with_bias=False, name='linear_node')(x)
    y = node_features + residual
    y = hk.LayerNorm(axis=-1, create_scale=True, create_offset=True,)(y)

    h = hk.nets.MLP(
        output_sizes=[NODE_MLP_HID]*NODE_MLP_NUM_LAYERS + [y.shape[-1]],
        activation=jax.nn.relu,
        name='mlp_node')(jnp.concatenate([y, global_features], axis=-1))
    h = h + y
    h = hk.LayerNorm(axis=-1, create_scale=True, create_offset=True,)(h)
    return h

  def update_global_fn(node_features, edge_features, global_features):
    x = jnp.concatenate([global_features, node_features], axis=-1)
    h = hk.nets.MLP(
        output_sizes=[GLOBAL_MLP_HID]*GLOBAL_MLP_NUM_LAYERS + [GLOBAL_EMBED_DIM],
        activation=jax.nn.relu,
        name='mlp_global')(x)
    h = h + global_features
    h = hk.LayerNorm(axis=-1, create_scale=True, create_offset=True,)(h)
    return h

  def attention_logit_fn(edge_features, sender_features, receiver_features, global_features):
    edge_features = edge_features[1]
    q = hk.Linear(NUM_HEADS * QK_DIM, with_bias=False, name='linear_query')(
        jnp.concatenate([sender_features, edge_features], axis=-1))
    q = jnp.reshape(q, list(q.shape[:-1]) + [NUM_HEADS, QK_DIM])
    k = hk.Linear(NUM_HEADS * QK_DIM, with_bias=False, name='linear_key')(receiver_features)
    k = jnp.reshape(k, list(k.shape[:-1]) + [NUM_HEADS, QK_DIM])
    return jnp.sum(q * k, axis=-1, keepdims=True) / (QK_DIM**0.5)

  def attention_reduce_fn(edge_features, weights):
    return edge_features[0] * weights

  def encode_graph(g):
    return jraph.GraphMapFeatures(
        embed_edge_fn=hk.Linear(EDGE_EMBED_DIM, name='linear_embed_edge'),
        embed_node_fn=hk.Linear(NODE_EMBED_DIM, name='linear_embed_node'),
        embed_global_fn=hk.Linear(GLOBAL_EMBED_DIM, name='linear_embed_global')
    )(g)

  @jax.vmap
  def _ppo_graph_part(g):
    g = encode_graph(g)
    enc_edges = g.edges

    for _ in range(NUM_GN_GAT_LAYERS):
      g = jraph.GraphNetGAT(
          update_edge_fn=update_edge_fn,
          update_node_fn=update_node_fn,
          attention_logit_fn=attention_logit_fn,
          attention_reduce_fn=attention_reduce_fn,
          update_global_fn=update_global_fn,
      )(g)
      g = g._replace(edges=enc_edges)

    return g

  def _ppo_graph_network(g):
    input_g = g
    input_globals = jnp.squeeze(g.globals, axis=1)
    input_globals = hk.Linear(GLOBAL_EMBED_DIM, with_bias=False)(input_globals)

    g = _ppo_graph_part(g)
    output_g = g
    g_globals = jnp.squeeze(g.globals, axis=1)
    g_nodes = g.nodes # B x n_nodes x NODE_DIM

    ######### The policy part
    # first predict an action distribution per block
    # for now lets try without further processing of the g_nodes
    node_feats = g_nodes

    BLOCK_ACT_DIM = 6
    block_act_locs = hk.Linear(
        BLOCK_ACT_DIM,
        w_init=hk.initializers.VarianceScaling(1e-4),
        b_init=hk.initializers.Constant(0.))(node_feats)
    block_act_scales = hk.Linear(
        BLOCK_ACT_DIM,
        w_init=hk.initializers.VarianceScaling(1e-4),
        b_init=hk.initializers.Constant(0.))(node_feats)
    block_act_scales = jax.nn.softplus(block_act_scales) + 1e-6

    # now for each gipper generate logits for which block it wants to move
    # for now lets try without further processing of g_globals
    global_feats = jnp.concatenate([input_globals, g_globals], axis=-1)

    NUM_GRIPS = 2
    grip_keys = hk.Linear(NUM_GRIPS * QK_DIM, with_bias=False)(global_feats)
    grip_keys = jnp.reshape(grip_keys, [grip_keys.shape[0], NUM_GRIPS, 1, QK_DIM]) # B x G x 1 x QK_DIM

    block_active = jnp.reshape(input_g.globals, [input_g.globals.shape[0], NUM_GRIPS, -1])
    block_active = block_active[:, :, -node_feats.shape[1]:]
    block_active = jnp.transpose(block_active, [0, 2, 1])
    block_keys = hk.Linear(QK_DIM, with_bias=False)(
        jnp.concatenate([node_feats, block_active], axis=-1)) # B x n_nodes x QK_DIM

    block_keys = jnp.reshape(block_keys, [block_keys.shape[0], 1, block_keys.shape[1], QK_DIM]) # B x 1 x n_nodes x QK_DIM

    grip_logits = jnp.sum(grip_keys * block_keys, axis=-1) / (QK_DIM**0.5) # B x G x n_nodes

    policy_output = PolicyOutput(
        block_act_locs=block_act_locs,
        block_act_scales=block_act_scales,
        grip_logits=grip_logits,)

    ######### The baseline value part
    # for now just doing it based on the globals

    # get the baseline value
    input_globals = jnp.squeeze(input_g.globals, axis=1)
    input_globals = hk.Linear(GLOBAL_EMBED_DIM, with_bias=False)(input_globals)
    g_globals = jnp.squeeze(output_g.globals, axis=1)
    g_globals = hk.Linear(GLOBAL_EMBED_DIM, with_bias=False)(g_globals)
    all_globals = jnp.concatenate([input_globals, g_globals], axis=-1)

    # trying something
    grip_logits_softmax = jax.lax.stop_gradient(jax.nn.softmax(grip_logits, axis=-1)) # dont want the value loss to influence gripper choice
    grip_logits_softmax = jnp.reshape(grip_logits_softmax, [grip_logits_softmax.shape[0], -1])
    all_globals = jnp.concatenate([all_globals, grip_logits_softmax], axis=-1)

    mlp_inputs = all_globals

    value_network = hk.nets.MLP(
        output_sizes=VALUE_MLP_LAYER_SIZES + [1],
        activation=jax.nn.relu)
    value = value_network(mlp_inputs)
    value = jnp.squeeze(value, axis=-1)

    return (policy_output, value)

  return _ppo_graph_network
\end{lstlisting}

%% file: full_size_plots.tex
\clearpage
\section{Additional Figures}
    \begin{figure}[h]
        \centering
        \includegraphics[width=\linewidth]{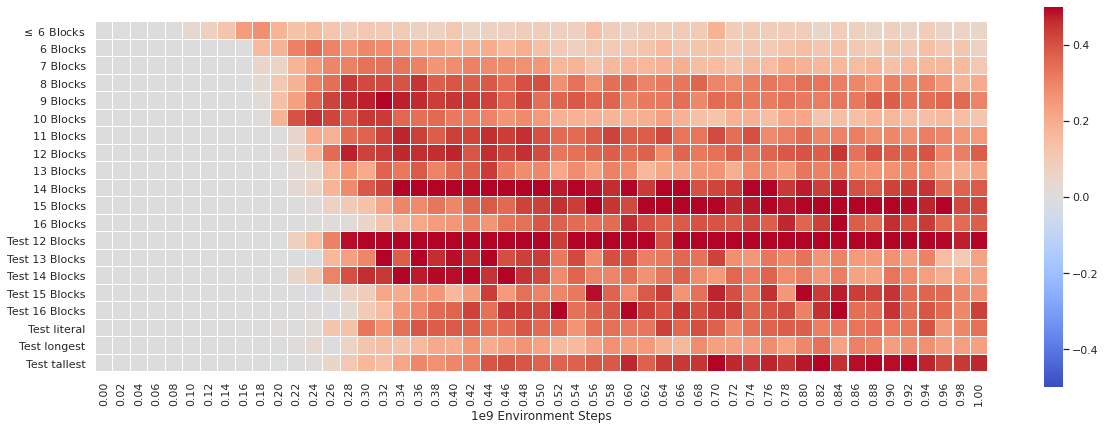}
        \caption{Plot showing \texttt{bimanual success rate} minus \texttt{single gripper success rate}. In our proposed magnetic assembly tasks, agents benefit strongly from having access to 2 grippers compared to 1. Full details in Figure \ref{fig:bimanual_full}.}
        \label{fig:bimanual_summary}
    \end{figure}
    
    \begin{figure}
        \centering
        \includegraphics[width=\linewidth]{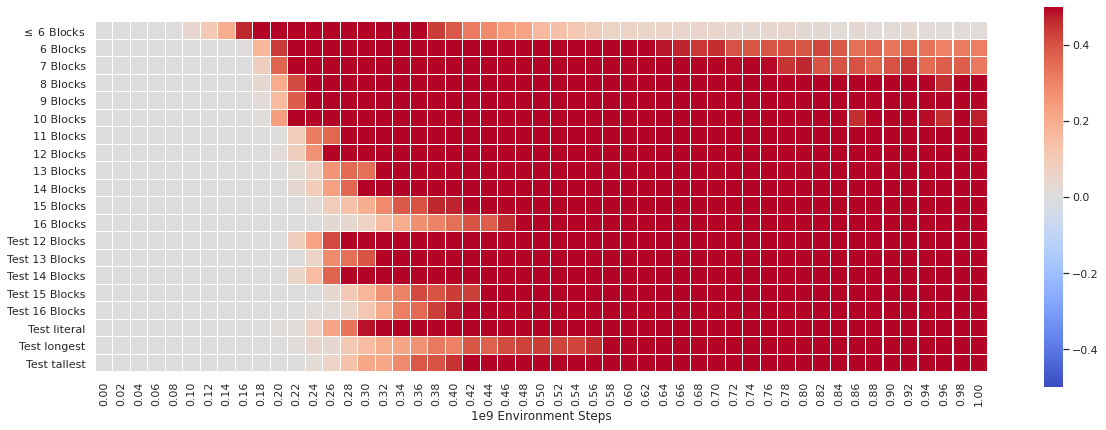}
        \caption{Plot showing \texttt{default agent success rate} minus \texttt{no attention agent success rate}, clearly demonstrating the necessity of the graph attention mechanism. Full details in Figure \ref{fig:no_attention_full}.}
        \label{fig:no_attention_summary}
    \end{figure}

\clearpage
\section{Full Size Version of All Plots}
    Due to the size of the plots, figures appear starting from the next page.
    \begin{figure}
        \centering
        \includegraphics[width=0.55\linewidth]{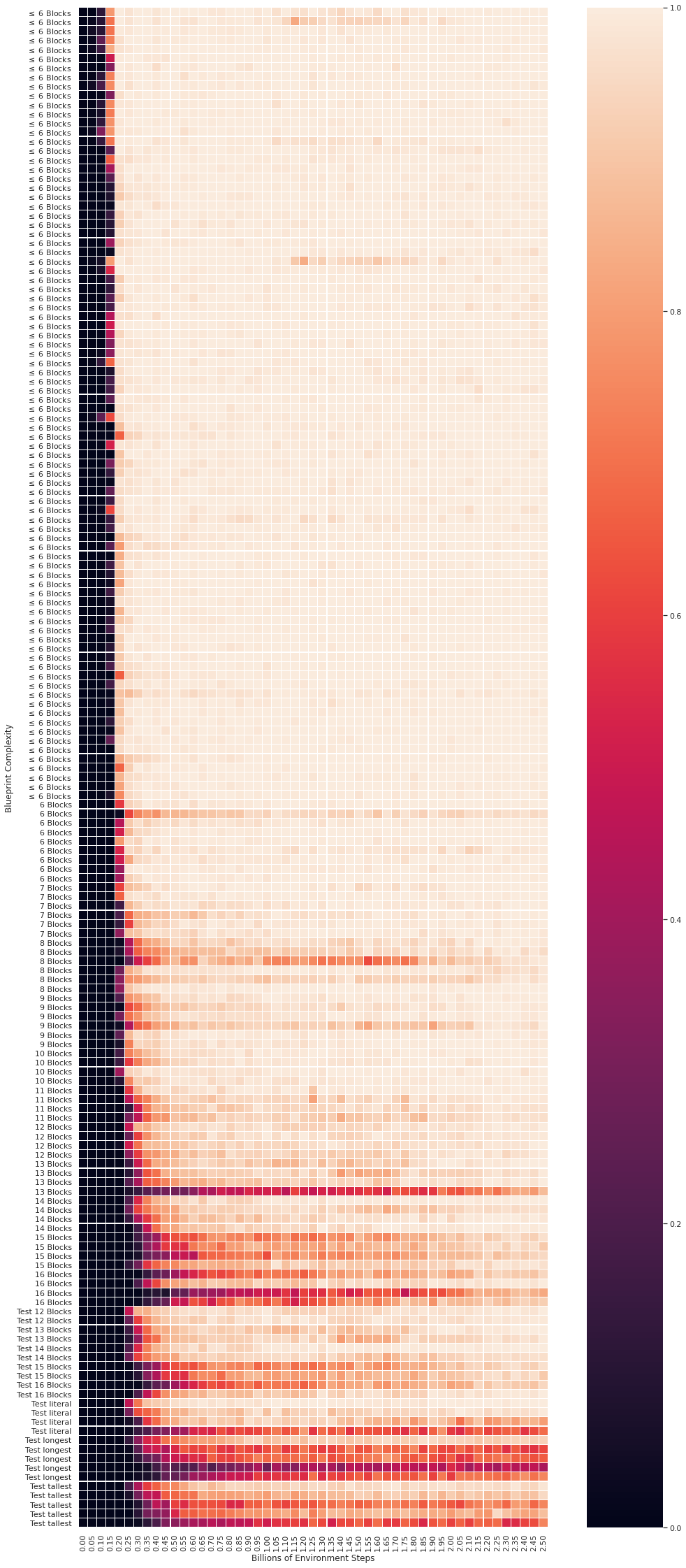}
        \caption{
        Plot presenting success rates for each blueprints throughout training, for our default agents (Section \ref{sec:agents}) and training procedure (Section \ref{sec:training}). Each square represents the success rate on 40 episodes, evaluated at that point in training, averaged across two training runs. Our results demonstrate (1) the key role of scale in training successful agents, and that (2) after a long period of training, agents generalize well to complex held-out blueprints.
        }
        \label{fig:super_policy_full}
    \end{figure}

    \begin{figure}
        \centering
        \includegraphics[width=0.55\linewidth]{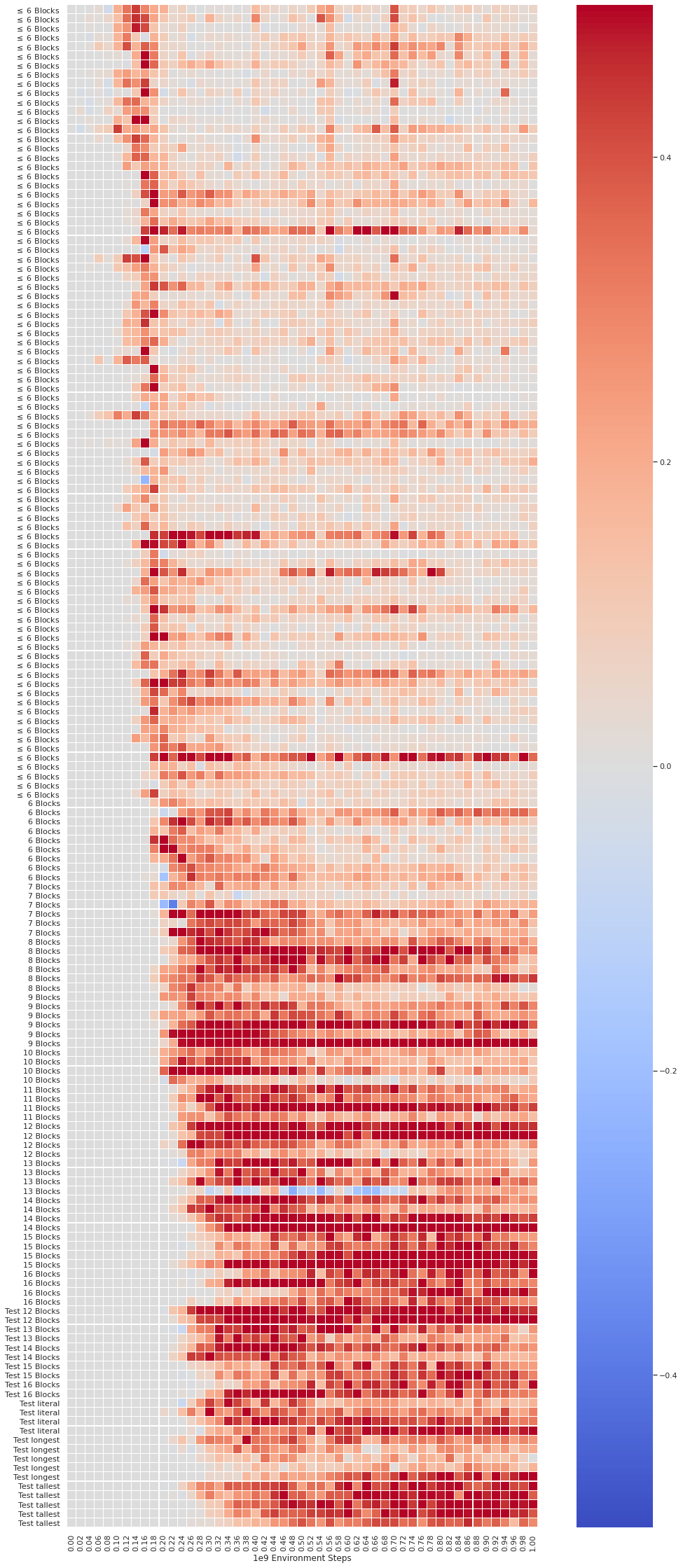}
        \caption{Plot showing \texttt{bimanual success rate} minus \texttt{single gripper success rate}. In our proposed magnetic assembly tasks, agents benefit strongly from having access to 2 grippers compared to 1.}
        \label{fig:bimanual_full}
    \end{figure}
    
    \begin{figure}
        \centering
        \includegraphics[width=0.55\linewidth]{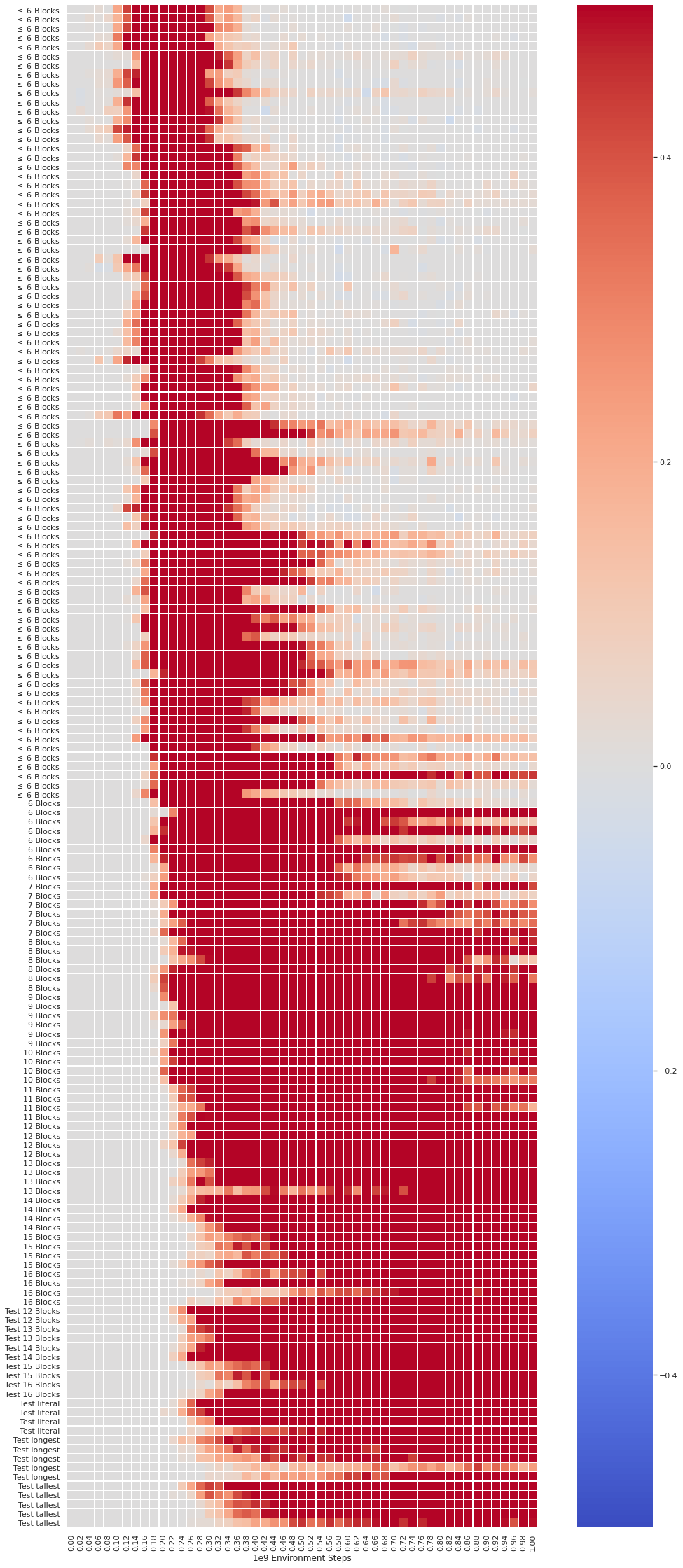}
        \caption{Plot showing \texttt{default agent success rate} minus \texttt{no attention agent success rate}, clearly demonstrating the necessity of the graph attention mechanism.}
        \label{fig:no_attention_full}
    \end{figure}

    \begin{figure}
        \centering
        \includegraphics[width=0.55\linewidth]{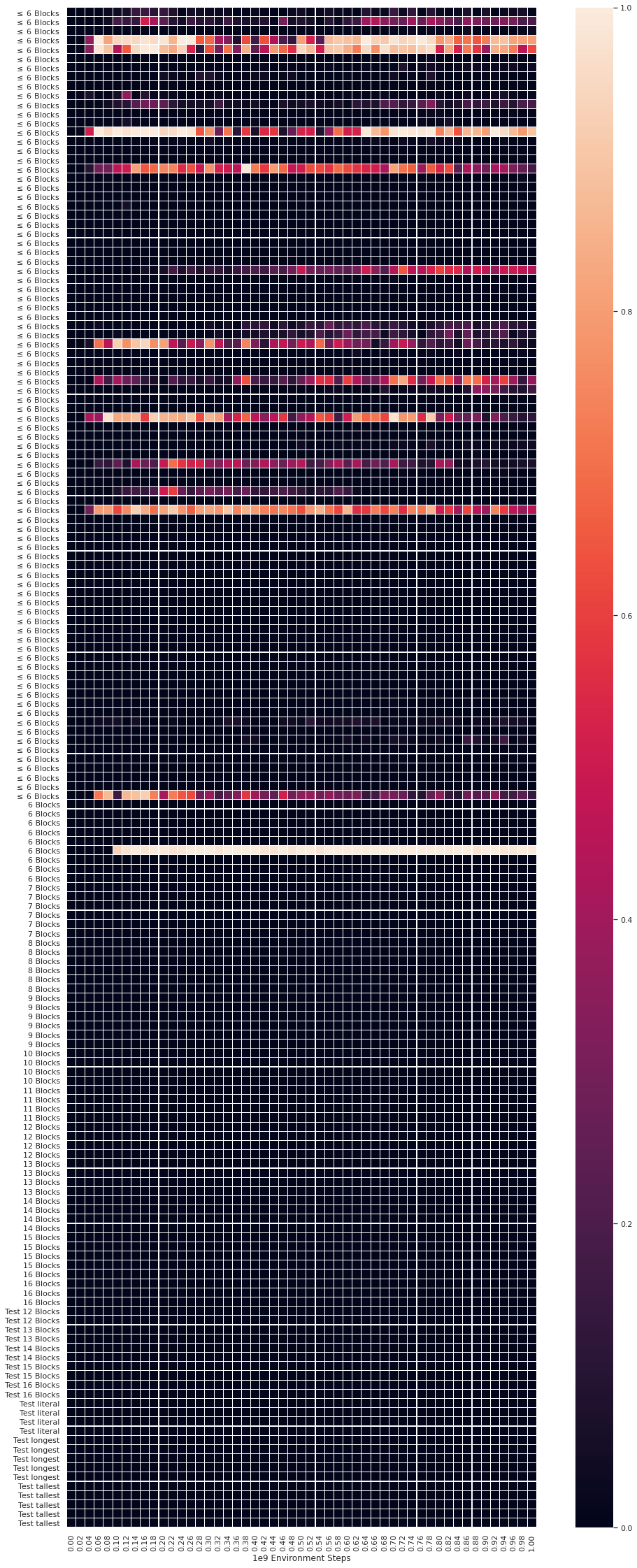}
        \caption{Plot showing success rates when training on only blueprint 6\_005.}
        \label{fig:just_6_005}
    \end{figure}
    
    \begin{figure}
        \centering
        \includegraphics[width=0.55\linewidth]{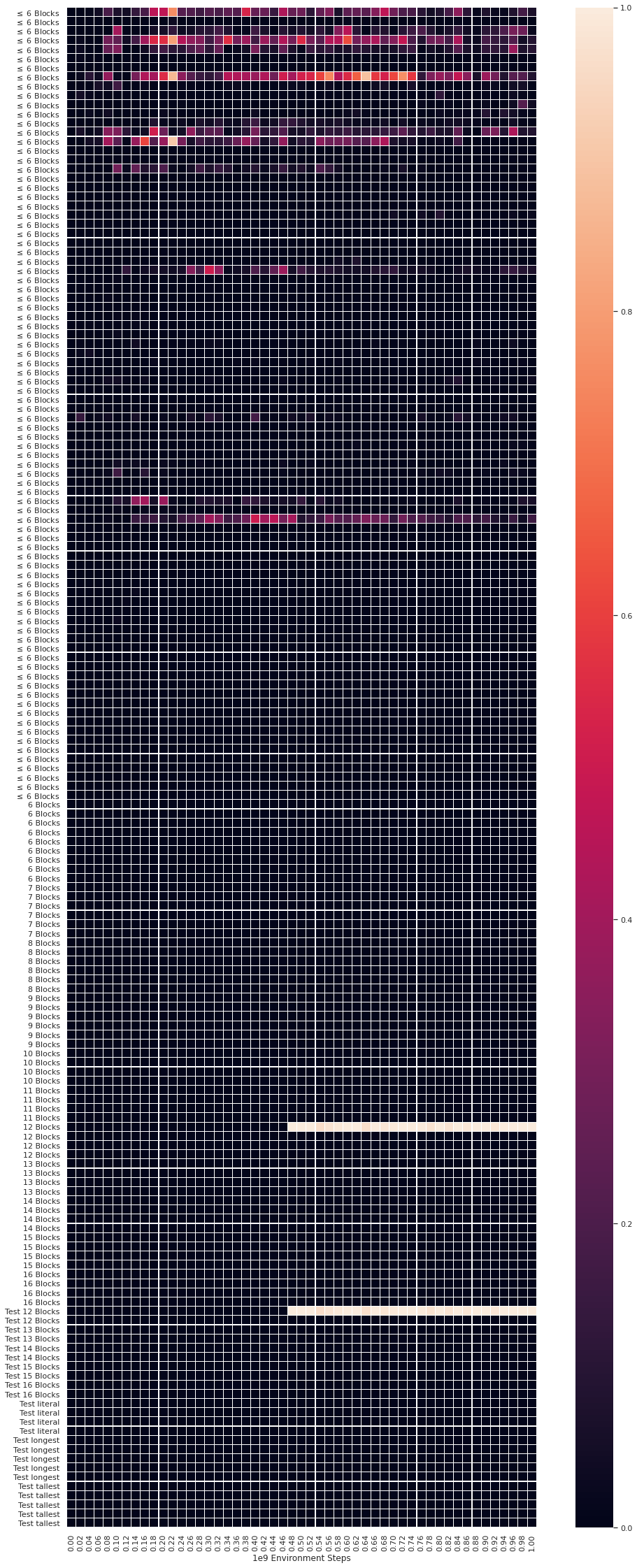}
        \caption{Plot showing success rates when training on only blueprint 12\_000.}
        \label{fig:just_12_000}
    \end{figure}
    
    \begin{figure}
        \centering
        \includegraphics[width=0.55\linewidth]{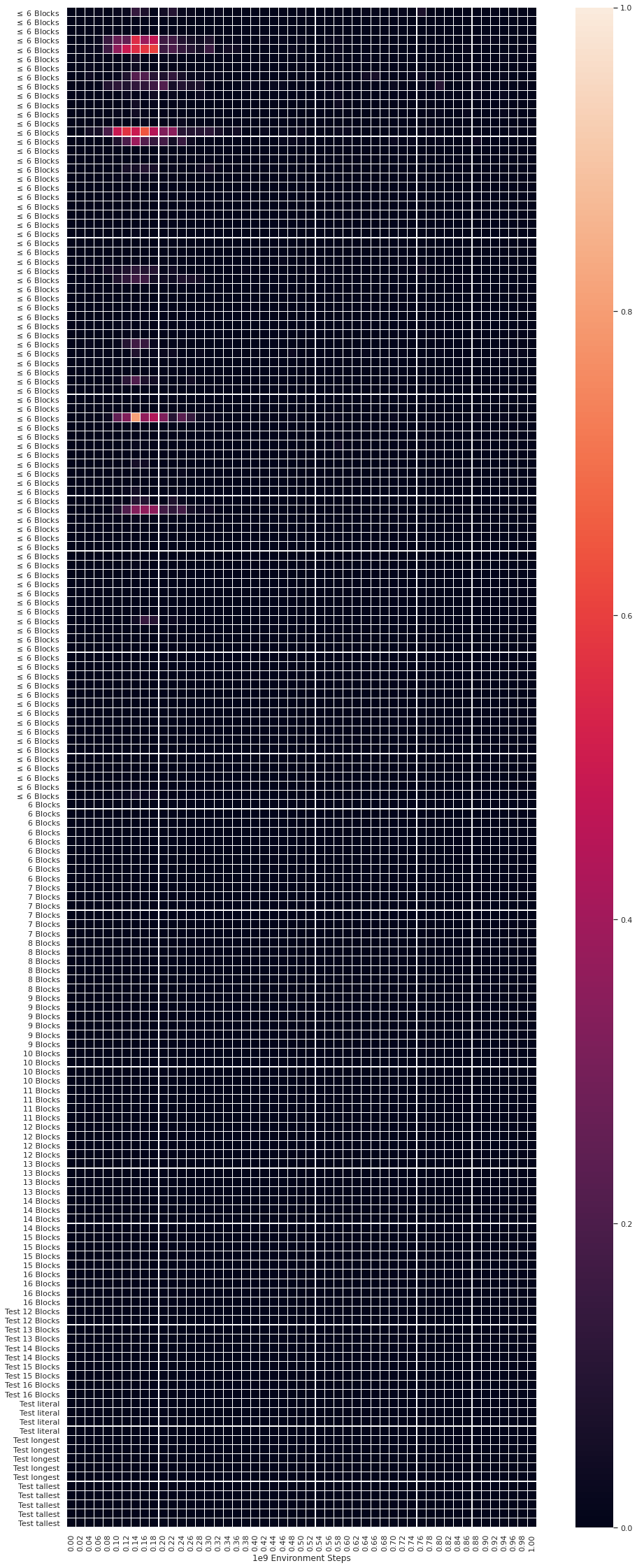}
        \caption{Plot showing success rates when training on only blueprint 16\_001.}
        \label{fig:just_16_001}
    \end{figure}

    \begin{figure}
        \centering
        \includegraphics[width=0.55\linewidth]{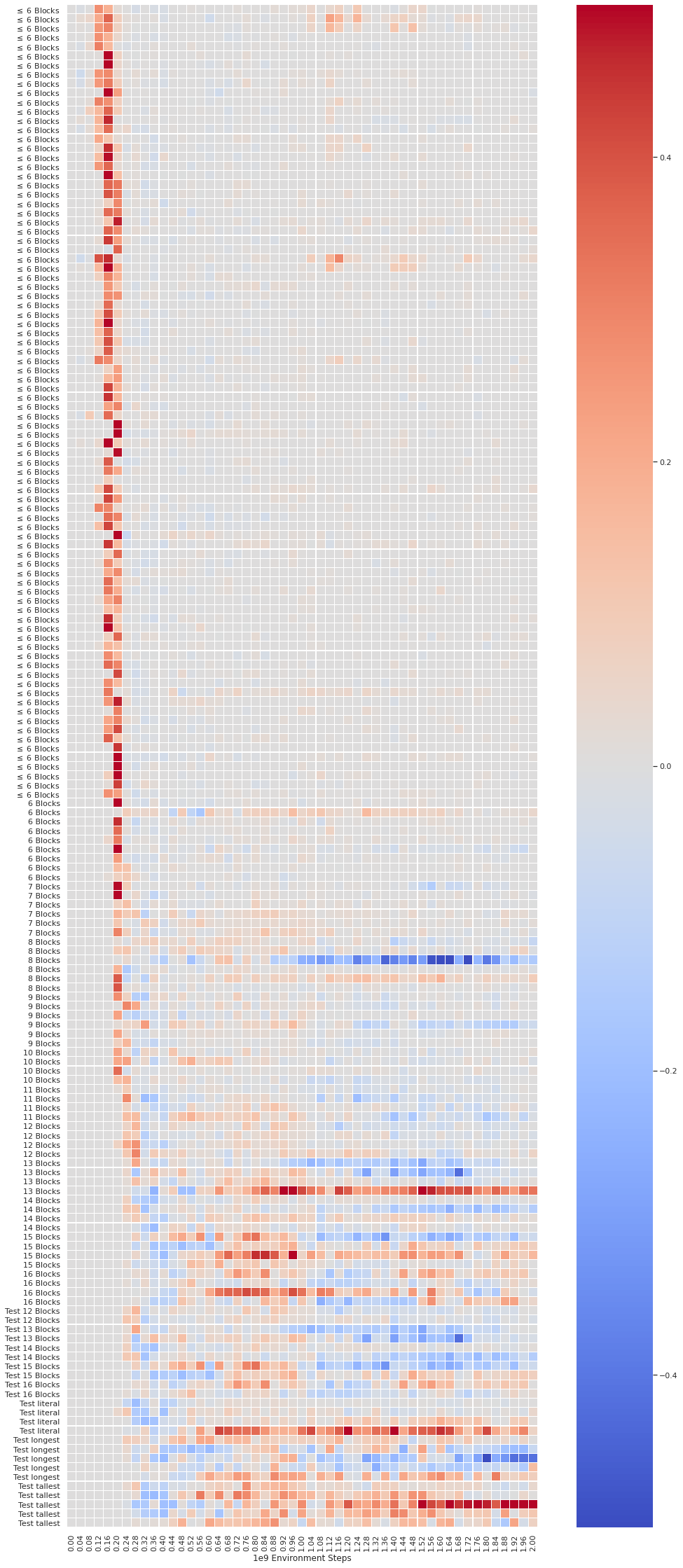}
        \caption{Plot showing \texttt{with curriculum agent (seed 0) success rate} minus \texttt{no curriculum agent success rate}.}
        \label{fig:curr_0_minus_no_curr}
    \end{figure}
    
    \begin{figure}
        \centering
        \includegraphics[width=0.55\linewidth]{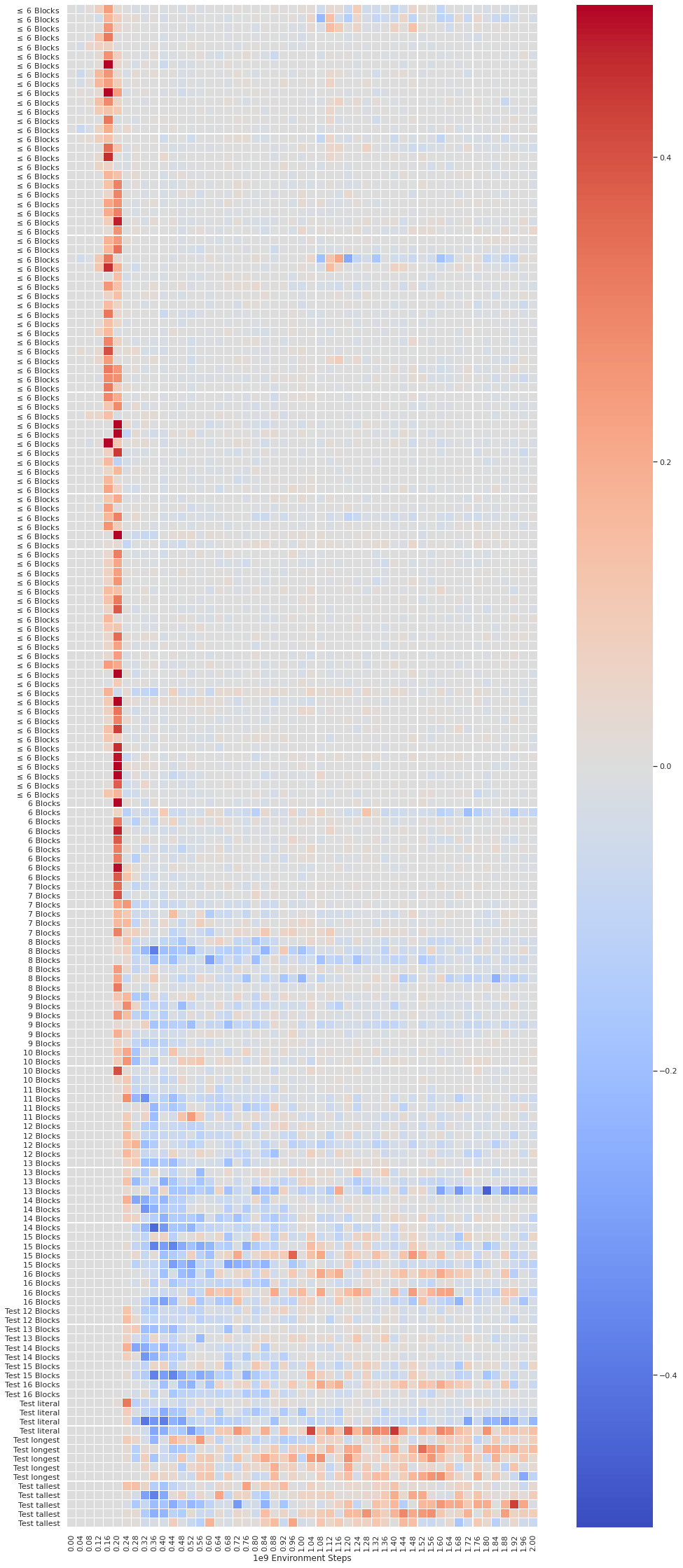}
        \caption{Plot showing \texttt{with curriculum agent (seed 1) success rate} minus \texttt{no curriculum agent success rate}.}
        \label{fig:curr_1_minus_no_curr}
    \end{figure}

    \begin{figure}
        \centering
        \includegraphics[width=0.55\linewidth]{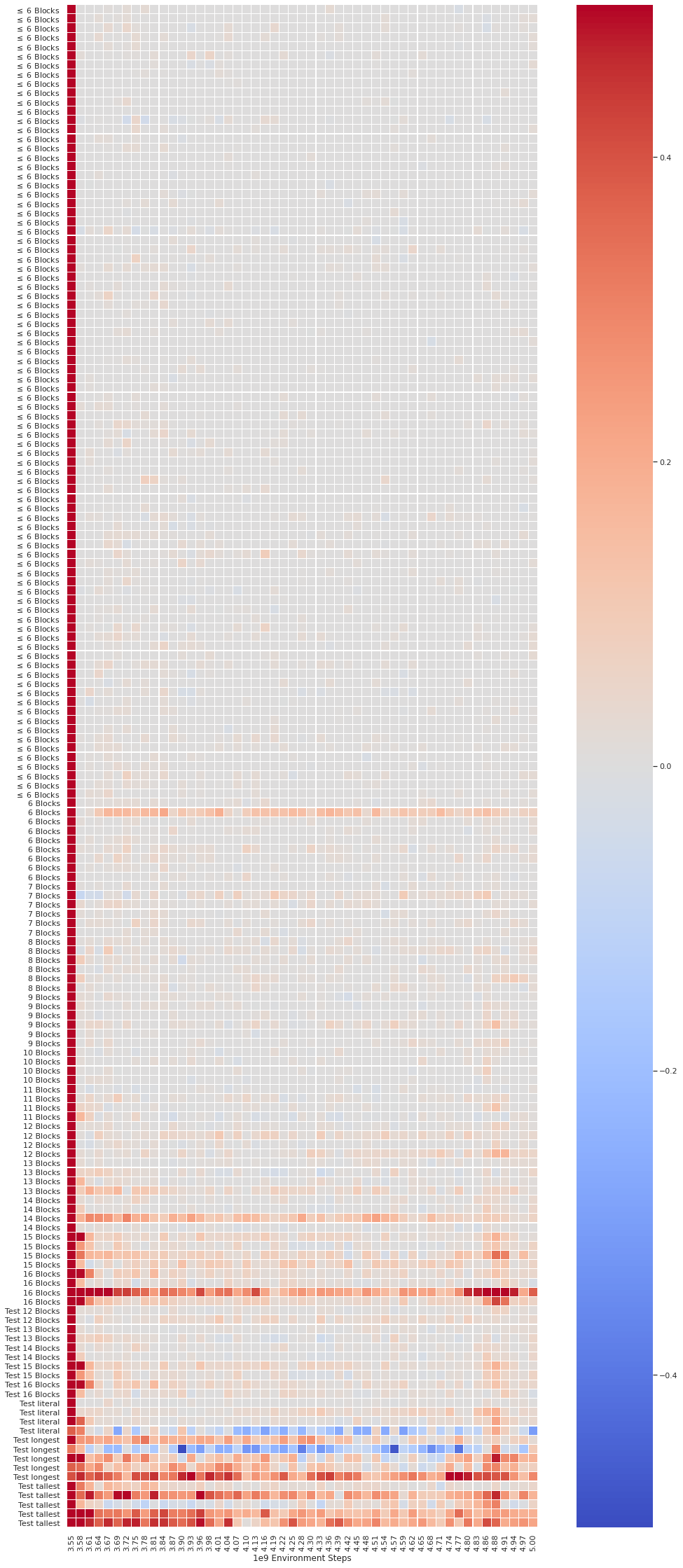}
        \caption{Plot showing \texttt{default agent success rate} minus \texttt{forked agent fine-tuned with gripper transition delay 2 success rate}.}
        \label{fig:super_minus_grip_trans_delay}
    \end{figure}

    \begin{figure}
        \centering
        \includegraphics[width=0.55\linewidth]{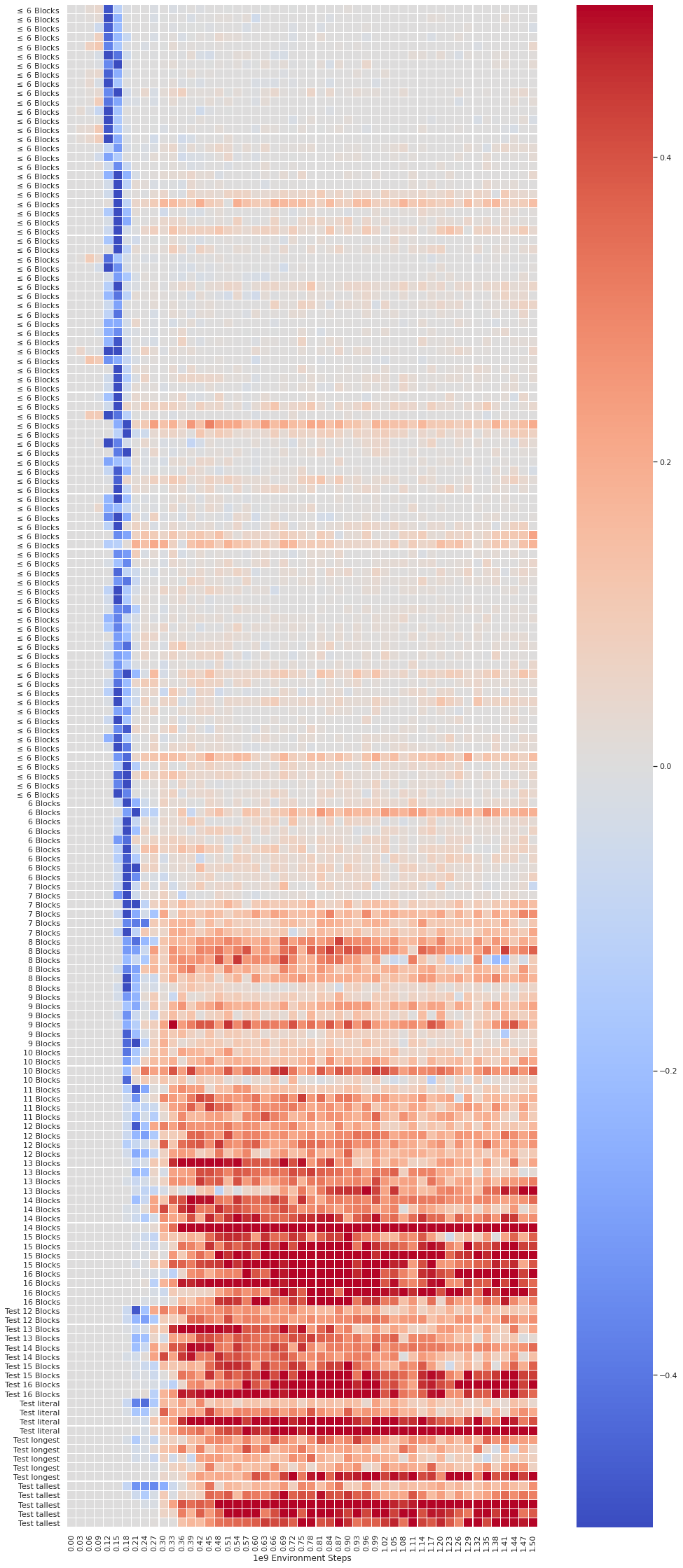}
        \caption{Plot showing \texttt{default agent success rate} minus \texttt{agent trained with gripper transition delay 2 from scratch success rate}.}
        \label{fig:super_minus_grip_trans_delay_from_scratch}
    \end{figure}
    
    \begin{figure}
        \centering
        \includegraphics[width=0.55\linewidth]{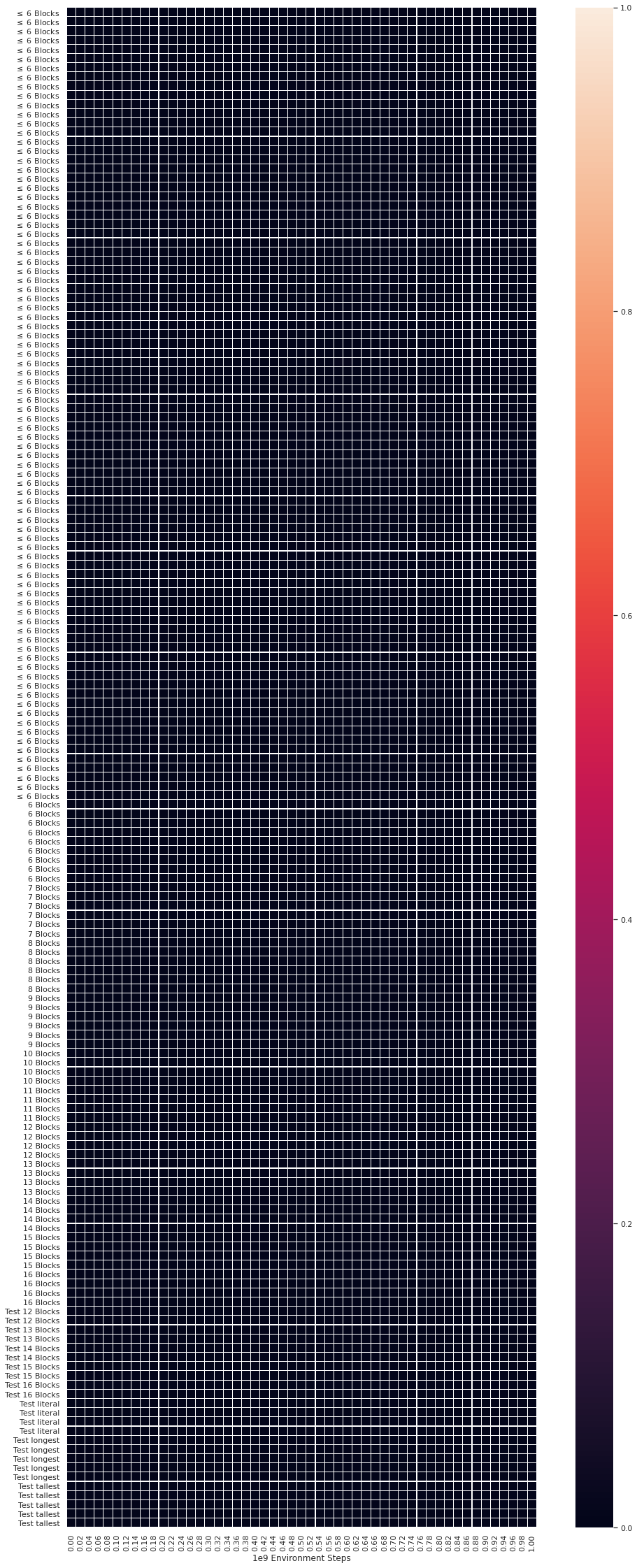}
        \caption{Plot showing the success rate of a residual network architecture trained on the full training set of blueprints.}
        \label{fig:resnet_all}
    \end{figure}
    
    \begin{figure}
        \centering
        \includegraphics[width=0.55\linewidth]{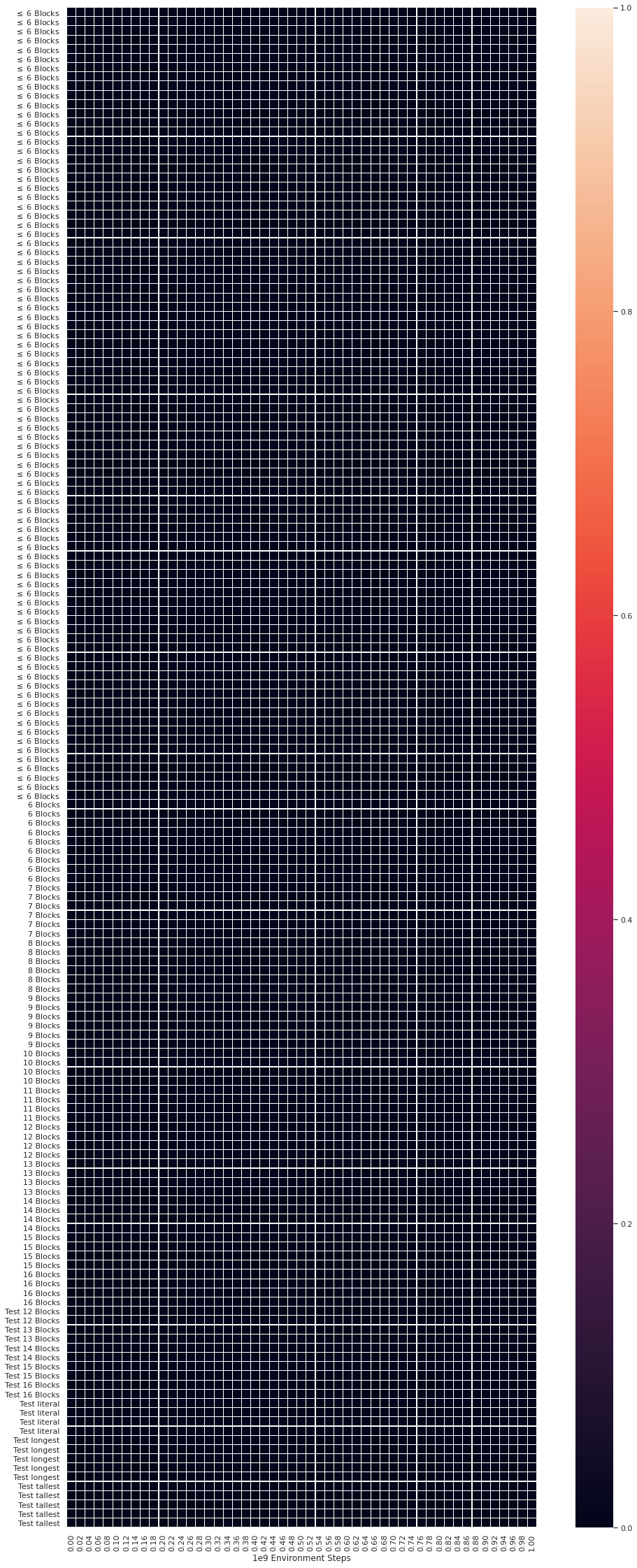}
        \caption{Plot showing the success rate of a residual network architecture trained on the subset of the training set of blueprints with $\leq 6$ blocks.}
        \label{fig:resnet_full}
    \end{figure}
    
    \begin{figure}
        \centering
        \includegraphics[width=0.55\linewidth]{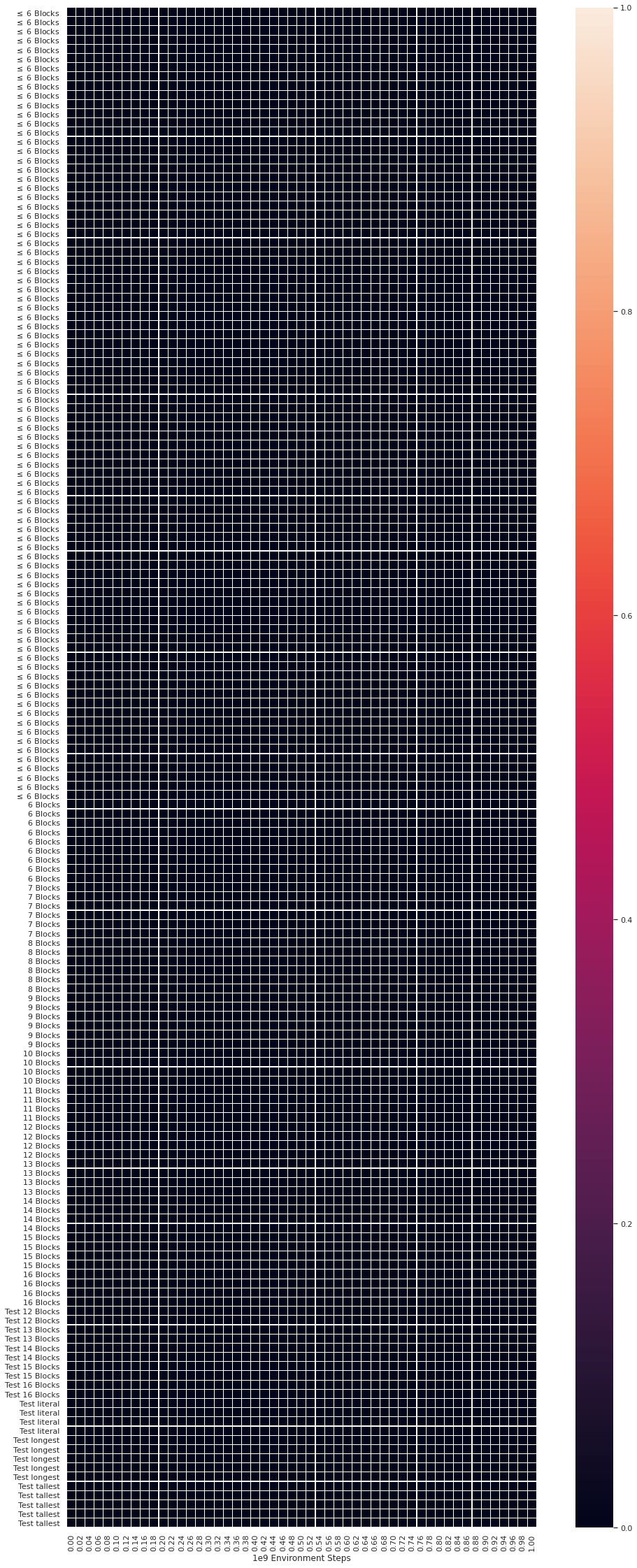}
        \caption{Plot showing the success rate of a residual network architecture trained only on a single blueprint requiring 6 blocks.}
        \label{fig:resnet_6_005}
    \end{figure}